\documentclass[10pt,twocolumn,letterpaper]{article}

\usepackage[pagenumbers]{cvpr} 

\usepackage{xcolor} 
\definecolor{LightBlue}{RGB}{173,216,230}
\usepackage{colortbl} 
\usepackage{graphicx}
\usepackage{booktabs}
\usepackage{multirow}
\usepackage{makecell}
\usepackage{arydshln}
\usepackage{rotating}
\usepackage{algorithm}
\usepackage{algpseudocode}
\usepackage{setspace}
\usepackage{amsmath}
\usepackage{bbm}
\usepackage{subcaption}
\pdfoutput=1

\definecolor{cvprblue}{rgb}{0.21,0.49,0.74}
\usepackage[pagebackref,breaklinks,colorlinks,allcolors=cvprblue]{hyperref}

\def\paperID{13775} 
\def\confName{CVPR}
\def\confYear{2025}

\title{Online Gaussian Test-Time Adaptation of Vision-Language Models}

\author{Clément Fuchs\thanks{\hspace{0.1cm} Equal contributions and corresponding authors. \texttt{\{clement.fuchs,maxime.zanella\}@uclouvain.be}} $^{\hspace{0.5mm} 1}$ \hspace{6mm} Maxime Zanella$^{*}$$^{1,2}$ \hspace{6mm} Christophe De Vleeschouwer$^{1}$
\\
$^{1}$UCLouvain, Belgium \hspace{6mm} $^{2}$UMons, Belgium \hspace{6mm} 
}

\begin{document}
\maketitle
\begin{abstract}
Online test-time adaptation (OTTA) of vision-language models (VLMs) has recently garnered increased attention to take advantage of data observed along a stream to improve future predictions. Unfortunately, existing methods rely on dataset-specific hyperparameters, significantly limiting their adaptability to unseen tasks. In response, we propose Online Gaussian Adaptation (OGA), a novel method that models the likelihoods of visual features using Gaussian distributions and incorporates zero-shot priors into an interpretable \textit{Maximum A Posteriori} (MAP) estimation framework with fixed hyper-parameters across all datasets. We demonstrate that OGA outperforms state-of-the-art methods on most datasets and runs. Additionally, we show that combining OTTA with popular few-shot techniques—a practical yet overlooked setting in prior research—is highly beneficial.
Furthermore, our experimental study reveals that common OTTA evaluation protocols, which average performance over at most three runs per dataset, are inadequate due to the substantial variability observed across runs for all OTTA methods.
Therefore, we advocate for more rigorous evaluation practices, including increasing the number of runs and considering additional quantitative metrics, such as our proposed Expected Tail Accuracy (ETA), calculated as the average accuracy in the worst 10\% of runs. 
We hope these contributions will encourage more rigorous and diverse evaluation practices in the OTTA community. Code is available at https://github.com/cfuchs2023/OGA. 
\end{abstract}

\begin{figure}[t]
\begin{center}
\includegraphics[width=0.5\textwidth]{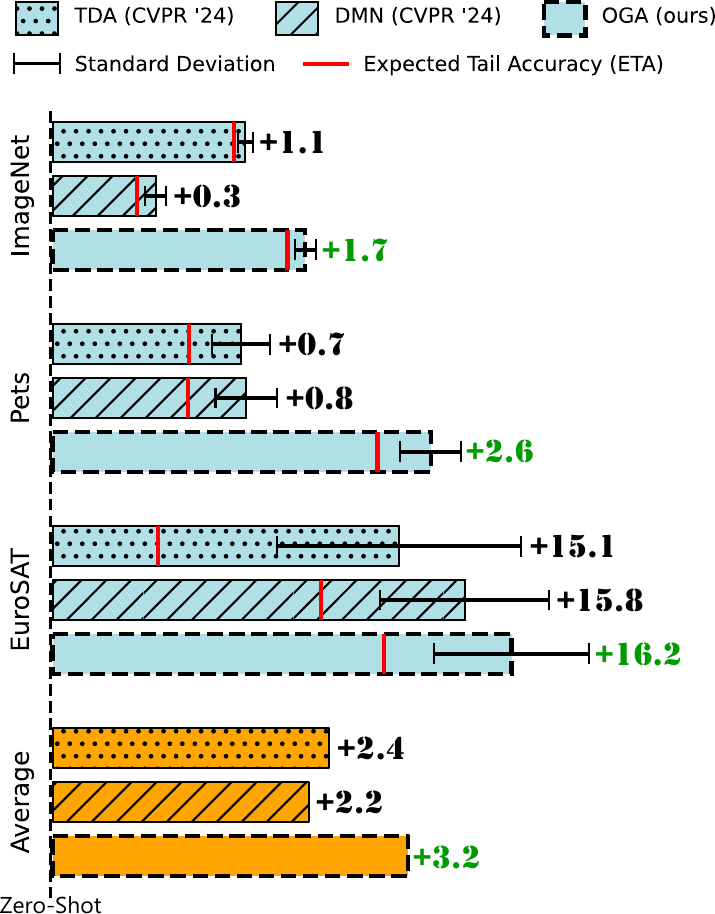}
\end{center}
\caption{The presented results are averaged over 100 runs. We propose the \textit{Expected Tail Accuracy} (ETA), i.e., the average over the 10\% worst runs, in solid red line. Our method named OGA not only significantly outperforms competitors on average but also has an ETA exceeding their average accuracy on several datasets (e.g., ImageNet and Pets). See Table \ref{tab:zero_shot_ACC_and_ETL} for more detailed results.}\label{fig:abstract_figure}
\end{figure}
\section{Introduction}
Vision-Language alignment has emerged as a powerful paradigm for pretraining models capable of handling a wide variety of downstream tasks with little or no labeled data. Contrastive methods such as CLIP \cite{radford_learning_2021} learn transferable visual representations by jointly optimizing a visual encoder and a textual encoder to align the representations of paired images and captions. This enables the creation of an image classifier without retraining the model, using textual descriptions of the classes. The classification procedure then relies simply on measuring the similarities between the textual features and those of the images, enabling zero-shot predictions. This has resulted in impressive zero-shot performance, as demonstrated on widely recognized supervised learning benchmarks such as ImageNet \cite{imagenet}. This success has motivated the investigation of methods to adapt vision-language models (VLMs) to unseen tasks, circumventing the need for training a model anew, either through prompt optimization \cite{zhou2022learning, shu_test-time_2022}, low-rank adaptation \cite{zanella2024low}, or adapters in the embedding space \cite{zhang_tip-adapter_2022, karmanov_efficient_2024}. These latter methods are of particular interest as they do not require access to the model weights—referred to as black-box methods \cite{ouali_black_2023, zanella_test-time_2024}—making them suitable for API-based applications.

Naturally, the test-time adaptation (TTA) paradigm has attracted considerable attention in the Vision-Language community, aiming at exploiting unlabeled data to further improve these performances. Notably, TTA has been deployed through methods which require encoding a large number of augmented views for each image \cite{shu_test-time_2022, zanella_test-time_2024} or rely on transductive settings that demand relatively large data batches to discover patterns among instances \cite{zanella_boosting_2024, martin_transductive_2024, stojnic_label_2024}. These limitations can be overcome when TTA is cast in an online setting, where data arrives in batches as small as one with the possibility of retaining information from one batch to the next ones. Very recent works, such as TDA \cite{karmanov_efficient_2024} and DMN \cite{zhang_dual_2024}, utilize cache models that are iteratively updated with incoming data. However, their performance depend strongly on some key hyper-parameters in their intricate prediction rule that must be adjusted specifically for each downstream task. This observation is not new and was recently highlighted in a study \cite{Silva-Rodriguez_2024_CVPR} on related cache-based methods \cite{zhang_tip-adapter_2022} in the few-shot setting. 
To mitigate this important practical deployment issue, we propose Online Gaussian Adaptation (OGA) which models the likelihoods of observed visual features with multivariate Gaussian distributions and combines them with the zero-shot priors, yielding a principled and interpretable \textit{Maximum A Posteriori} (MAP) prediction rule (with no need for \textit{hyper-parameters} tuning). Our approach achieves superior performances, as depicted in Figure \ref{fig:abstract_figure} and Table \ref{tab:zero_shot_ACC_and_ETL}.

Additionally, our study reveals that, despite their growing popularity, online test-time adaptation (OTTA) methods for VLMs lack rigorous and relevant evaluation frameworks. For instance, TDA \cite{karmanov_efficient_2024} and DMN \cite{zhang_dual_2024} evaluate performance using at most three random seeds, even though Figure \ref{fig:abstract_figure} and Table \ref{tab:zero_shot_ACC_and_ETL}
 demonstrate significant variance in measured accuracy across random runs. We propose measuring the average accuracy over more runs to mitigate variability in comparisons arising from the stochastic nature of data stream generation. 
 Furthermore, we argue that the average accuracy metric is insufficient to accurately compare methods, as it fails to account for \textit{tail risk}, where methods may exhibit significantly worse accuracies for a small proportion of runs. This behavior could render a method undesirable in practice. Therefore, we recommend reporting an additional metric, which we term \textit{Expected Tail Accuracy} (ETA). ETA represents the average accuracy below the lower $10^{\text{th}}$-percentile, capturing performance in worst-case scenarios.
\paragraph{Contributions.} We summarize our contributions as follows: 
\begin{itemize}
    \item We propose Online Gaussian Adaptation (OGA), an OTTA method that models the likelihoods of observed visual features with multivariate Gaussian distributions and combines them with zero-shot priors into an elegant \textit{Maximum A Posteriori} (MAP) prediction rule with \textit{fixed hyper-parameters} across all datasets. Our method delivers strong performance, fits in the blackbox framework, and is computationally efficient. 
    \item Similar to previous works, we report performances of OTTA methods when applied to zero-shot VLMs. OGA outperforms state-of-the-art methods on most datasets and runs. Additionally, we compare methods atop popular few-shot methods, a very convenient way to combine offline few-shot learning with efficient online adaptation which has been overlooked so far in OTTA.
     \item Finally, we advocate for more rigorous evaluation procedures in this domain, emphasizing the need for multiple runs to account for variability and introducing \textit{Expected Tail Accuracy} (ETA) as a metric to assess performance in worst-case scenarios.
\end{itemize}

\section{Related work}
\label{sec:related_work}

\paragraph{Fine-tuning of VLMs.} One main design choice that differentiates fine-tuning methods is the set of parameters they tune, from input textual tokens \cite{zhou2022learning, shu_test-time_2022, huang2022unsupervised, ma2024swapprompt}, hidden layers \cite{zanella2024low}, additional parameters at the output of the text or vision encoder \cite{yu2023task, gao2024clip}, adapters as memory banks \cite{zhang_tip-adapter_2022, karmanov_efficient_2024, zhang_dual_2024}. Others operate directly in the embedding space, for example with a mode-seeking algorithm \cite{zanella_test-time_2024}. One notable group of such methods, sometimes called black-box methods in the literature \cite{ouali_black_2023, zanella_test-time_2024}, is undoubtedly cache-based methods. These methods stem from the initial work of Tip-Adapter \cite{zhang_tip-adapter_2022}, which explicitly combines logits from zero-shot prediction with similarity scores derived from a memory bank. Other notable advances in black-box methods include the recent successes of Gaussian modeling in few-shot learning \cite{wang_hard--beat_2024} and in transductive settings \cite{zanella_boosting_2024}. Both approaches refine class representations directly within the embedding space, modeling them as a balanced mixture of multivariate Gaussian distributions. Inspired by these recent developments in related fields, we propose to model the likelihoods of observed visual features with multivariate Gaussian distributions. We  then use the resulting posterior probabilities obtained from these likelihoods and the zero-shot priors to yield a principled \textit{Maximum A Posteriori} (MAP) prediction rule that is both interpretable and mathematically sound.

\paragraph{Test-Time Adaptation of VLMs.} The major distinction between current TTA methods lies in how they process the incoming data. One group of methods operates on a single image with data augmentations at test time, such as TPT \cite{shu_test-time_2022} which relies on prompt optimization for each individual image. MTA \cite{zanella_test-time_2024} avoids prompt tuning and optimizes a mean-shift-inspired objective function. However, these methods substantially increase computational requirements. In transductive learning, another branch of unsupervised learning, VLMs are directly adapted to the testing data. For example, EM-Dirichlet \cite{martin_transductive_2024} optimizes a maximum likelihood estimator of a Dirichlet distribution directly in the prediction space. ZLaP \cite{stojnic_label_2024} proposes propagating zero-shot labels based on a similarity graph of the representation of each instance. TransCLIP \cite{zanella_boosting_2024, zanella2024boostinghisto, khoury2024enhancing} suggests adding a text-based regularization derived from a Kullback-Leibler divergence term in an expectation-maximization-like objective function. One major drawback of these methods is that they rely on relatively large batch sizes, and require multiple samples of the same class within a batch to effectively leverage relationships between instances.

\paragraph{Online Test-Time Adaptation of VLMs.} OTTA approaches treat incoming data as a stream, retaining information from one batch to the next ones. A nascent work is \cite{ma2024swapprompt}, although it does not fit in the blackbox framework and uses a computationally expensive strategy combining prompt tuning and augmentations. 
More recent works provide a highly efficient solution to these issues by maintaining a small cache of selected samples to iteratively improve a prediction rule. Notable examples are TDA \cite{karmanov_efficient_2024} and DMN \cite{zhang_dual_2024}, which both use a similar minimal-entropy filtering strategy to fill their cache and a prediction rule directly inspired by Tip-Adapter. However, these methods rely on hyper-parameters that are difficult to tune for each new benchmark. In contrast, our approach is simple and practical, using just one interpretable hyper-parameter to weight the learned likelihoods.




\section{Preliminaries}
\label{sec:preliminaries}

To understand recent adaptation methods for vision-language models (VLMs), we start by defining the core components of the classification pipeline. At its foundation, a VLM encodes both images and textual descriptions into a shared embedding space, enabling comparison and alignment. These descriptions are tokenized into textual inputs ${\mathbf c}_k$, where $1<k\leq K$ ($K$ the number of classes), which are then transformed by the textual encoder into normalized embeddings ${\mathbf t}_k$ on a unit-hypersphere. The image ${\mathbf x}_i$, where $i = 1, \dots, N$, is processed by the visual encoder to produce embeddings ${\mathbf f}_i \in \mathbb{R}^d$, where $d$ is the dimension of the embedding space. These embeddings are also normalized to lie on the unit-hypersphere, facilitating direct comparison between images and class descriptions. With this shared embedding space, the cosine similarity between textual and visual representations ${\mathbf f}_i^\top {\mathbf t}_k$ forms the basis for classification tasks.

\paragraph{Zero-shot prediction.} Deploying VLMs in a zero-shot setting is one of the simplest and most direct ways to perform downstream tasks, leveraging the pre-training process described in \cite{radford_learning_2021}. To classify an image, the similarity between the image embedding and each class embedding is measured using cosine similarity, producing logit scores 
\begin{equation} \label{logits-ik} l_{i,k} = {\mathbf f}_i^\top {\mathbf t}_k. \end{equation}
These logits can be transformed into probabilistic predictions through a softmax function, which computes the posterior probability of class $k$ given the test image ${\mathbf x}_i$     
\begin{equation}
\label{zero-shot-prediction}
y_{i,k}= \frac{\exp (l_{i,k}/\tau)}{\sum_{j}^K \exp (l_{i,j}/\tau)}  
\end{equation}
 where $\tau$ is the softmax temperature parameter that controls the sharpness of the probability distribution.  The image ${\mathbf x}_i$ can then be classified by selecting the class with the highest posterior probability: $\hat{k} = \mathrm{argmax}_{k}~y_{i,k}$.


\paragraph{Few-shot adaptation.} When few shots are available, they can be used to learn richer representations of the classes in the textual embedding space. This is done either (i) by fine-tuning the input prompts (so as to minimize the cross-entropy loss computed on the few available shots), as in prompt-tuning methods like CoOp~\cite{zhou2022learning}; or (ii) by updating a set of additional parameters called adapters ~\cite{zhang_tip-adapter_2022} typically directly at the output of the model such as TaskRes \cite{yu2023task}. Respectively, we have:
\begin{equation}
    {\mathbf c}_k^{\text{CoOp}}=(\mathbf{v}_k^1, \dots, {\mathbf v}_k^M, [\mbox{class}_k]); \, \quad {\mathbf t}_k^{\text{TaskRes}} = {\mathbf t}_k + \alpha {\mathbf b}_k
    \label{eq:coop_and_taskres}
\end{equation}
where $( {\mathbf v}_k^l)_{1 \leq l \leq M}$ are trainable text tokens, $[\mbox{class}_k]$ is the fixed class tokens, ${\mathbf b}_k$ class-wise learnable parameters, and $\alpha$ a scaling hyper-parameter. Observe that prompt tuning incur heavy computational load for fine-tuning and might be hard to optimize, since every gradient update of the text input requires back-propagating through the entire model\footnote{We refer to the runtime studies of \cite{zanella_test-time_2024, karmanov_efficient_2024}.}.
Note that our method is orthogonal to those advances in the few-shot learning community, in fact we show that our proposed OGA and other OTTA methods can be applied atop of them (see Table \ref{tab:atop} with CoOp and TaskRes), offering a very convenient approach where few-shot supervised learning is done offline (potentially with heavy computation) with further adaptation done online using an efficient OTTA method.

\paragraph{Cache model.}
One of the first works to use a cache for VLMs adaptation is Tip-Adapter \cite{zhang_tip-adapter_2022}, which stores few-shot samples. In its training-free version, it directly utilizes the cache for final predictions by combining zero-shot similarities with cache similarities to compute adapted logits,
\begin{equation}
l_{i,k} = \mathbf f_i^T {\mathbf t}_k +  \alpha  \sum_{m} \exp(-\beta (1 - {\mathbf f}_i^T \mathbf{f}_{m}^{(k)}))
\label{eq:tip-adapter}
\end{equation}
with $\mathbf{f}_{m}^{(k)} \in \mathbb{R}^d$ the $m^{\text{th}}$ sample held in the cache for the $k^{\text{th}}$ class , $\alpha$ and $\beta$ being hyper-parameters. This adaptation function was later used in an online setting by TDA \cite{karmanov_efficient_2024}. Note that, unlike Tip-Adapter, TDA relies on pseudo-labels rather than ground truth labels, as it focuses on zero-shot adaptation. A major drawback of these Tip-Adapter-based methods is their dependence on key hyper-parameters ($\alpha$ and $\beta$) that must be carefully tuned for each downstream task \cite{Silva-Rodriguez_2024_CVPR}. This is is done via intensive searches over validation sets, requiring additional labeled samples which reduces their portability to new tasks. Our OGA method addresses this limitation with a principled MAP prediction rule, as explained in the next section.


\section{Online Gaussian Adaptation}

This section introduces our proposal to improve the zero-shot capabilities of a pre-trained VLM, based on the knowledge captured by a set of samples whose class is known with high confidence.
In an online setting, those samples are continuously collected along the stream, to fill in and then update a cache memory. In practice, we select the samples with the smallest zero-shot prediction entropy, i.e. those reliably labeled by the zero-shot classifier. The selected samples are then used to estimate a model of the image features class-conditional likelihoods as multivariate Gaussian distributions. The likelihoods are subsequently combined with the zero-shot prediction, considered as a prior, to estimate the class posterior for a new sample, using a prediction rule derived from Bayes formula. The main steps involved in this process—namely class posterior estimation, Gaussian parameters estimation, and online selection of reliable samples—are detailed below.

\paragraph{Gaussian modeling.}
Gaussian Mixture Models (GMM) have been succesfully used for both zero-shot and few-shot adaptation of VLMs \cite{wang_hard--beat_2024, zanella_boosting_2024}. We adopt this framework to model the image feature likelihoods conditioned on the class. Hence, for the feature $\mathbf{f}_i$ associated to image $i$, we have $ 
p_{i,k} = p(\mathbf{f}_i | c_i = k) = p(\mathbf{f}_i | \boldsymbol{\mu}_k, \Sigma, k),
$
following a multivariate normal distributions with shared covariance $\Sigma$. Formally,
\begin{equation}
p_{i,k} = \dfrac{1}{2\sqrt{(2\pi^d |\Sigma|}} \exp(-\dfrac{1}{2} (\mathbf{f}_i - \boldsymbol{\mu}_k)^T P(\Sigma) (\mathbf{f}_i - \boldsymbol{\mu}_k))
\label{eq:gaussian-pdf}
\end{equation} where $P(\Sigma)$ is an estimator of the precision matrix $\Sigma^{-1}$.
\paragraph{Pseudo-Bayesian adaptation rule.}
Our proposed adaptation rule is derived from the class posterior probabilities given by the Bayes rule. This posterior reads as 
\begin{equation}
p(c_i = k | \mathbf{f}_i) = \dfrac{p_{i,k} \cdot p(c_i = k)}{p(\mathbf{f}_i)} = \dfrac{p_{i,k} \cdot p(c_i = k)}{\sum_{l=1}^{K} p_{i,l} \cdot p(c_i = l)}.
\label{eq:bayes-rule}
\end{equation} In absence of prior knowledge about class probability, the prior $p(c_i = k)$ is generally chosen as $1/K$ to model the features distribution as a balanced mixture of multivariate normals. However, in the case of VLMs, we propose to leverage the knowledge obtained from the zero-shot predictions by using the soft labels $y_{i,k}$ as priors, which yields 
\begin{equation}
\label{eq:map-prediction-rule}
p(c_i = k | f_i) = \dfrac{p_{i,k} \cdot y_{i,k}}{\sum_{l=1}^{K} p_{i,l} \cdot y_{i,l}}.
\end{equation}
Interestingly, one could remark that Eq. \eqref{eq:map-prediction-rule} yields a \textit{Maximum A Posteriori} (MAP) estimator for each of the sample. To better control the degree to which the initial zero-shot prediction is modified by the Gaussian likelihoods, we introduce an hyper-parameter $\nu$
\begin{equation}
\label{eq:map-prediction-rule-tau}
    p(c_i = k | f_i) = \dfrac{p_{i,k}^{\nu} \cdot y_{i,k}}{\sum_{l=1}^{K} p_{i,l}^{\nu} \cdot y_{i,l}}.
\end{equation}
We use the same \textit{fixed} value of $\nu = 0.05$ across all datasets, and investigate its impact in our ablation study (see Figure~\ref{fig:ablation_tau}).
\paragraph{Gaussian parameters update.}
Whenever the cache memory is updated, we also update the Gaussian parameters. First, the centroids $\boldsymbol{\mu}_k$ are updated as the mean of the cached samples for the $k^{\text{th}}$ class. Then, the shared covariance matrix is updated using the cached samples as 
\begin{equation}
\Sigma = \dfrac{1}{n-1} \sum\limits_{k=1}^{K} \sum\limits_{m} (\mathbf{f}_{m}^{(k)} - \boldsymbol{\mu}_k) (\mathbf{f}_{m}^{(k)} - \boldsymbol{\mu}_k)^T
\label{eq:sigma-update}
\end{equation} where $n$ is the total number of samples in the cache and $\mathbf{f}_{m}^{(k)}$ the $m^{\text{th}}$ cached sample for class $k$. 
Note that since we store a relatively low (typically at most $8$) number of samples per class, the total number of samples used for estimating $\Sigma$ can be lower or on the same order of magnitude as the embedding space dimension $d$. Therefore, in the case where we have less than $4d$ samples in our cache, we use the Bayes-Ridge estimator of \cite{kubokawa_estimation_2008} which reads as 
\begin{equation}
P = d (n_t \Sigma + tr(\Sigma) I_{d})^{-1}.
\label{eq:bayes-ridge-estimator}
\end{equation}
When more than $4d$ samples are in the cache, we revert to using the inverse of $\Sigma$ as $P(\Sigma)$. More details are provided in the ablation study in Table~\ref{tab:ablation_P}.

\paragraph{Online selection of samples.}
Similarly to~\cite{karmanov_efficient_2024}, the samples are selected to fill in the cache according to their zero-shot entropy. More specifically, we compute the zero-shot Shannon entropy for a single sample from its zero-shot soft labels as $e_i = -\sum_{k = 1}^{K} \log(y_{i,k}) y_{i,k}$.  If the sample's entropy is lower than that of at least one cached sample for the class matching its pseudo-label, we replace the cached sample with the highest entropy with this new one. This process builds a low-entropy cache for each class as the model encounters new data.
\section{Experimental setting}
\paragraph{Datasets.}
We follow the settings of previous works \cite{zhou2022learning} and use ImageNet \cite{imagenet} as well as 10 other datasets: SUN397 \cite{sun397} for fine grained classification of scenes,  Aicraft \cite{aircraft} for classification of aircraft types, EuroSAT \cite{eurosat} for satellite imagery, StanfordCars \cite{cars} for cars models, Food101\cite{food} for food items, Pets \cite{pets} for pet types, Flower102 \cite{flower} for flowers species, Caltech101 \cite{caltech101} for a variety of general objects, DTD \cite{dtd} for textures types and UCF101 \cite{ucf101} for actions recognition.
\paragraph{Zero-shot model.}
We use CLIP with a ViT-B/16 visual architecture for all experiments.
\paragraph{Data stream generation.}
We generate i.i.d. data streams from the test set of each dataset, and then run the methods on the full stream with batch size $32$. For each dataset, the methods are compared on the \textit{same} 100 runs. In our ablation study, we provide further results for our approach for batch sizes $1$, $64$ and $128$ in Table~\ref{tab:ablation_bs}.

\paragraph{Competitors.}
We compare our approach to two recent state-of-the-art works in OTTA, namely TDA (CVPR '24) \cite{karmanov_efficient_2024} and DMN (CVPR '24) \cite{zhang_dual_2024}.  For the sake of fairness, we use the same total cache size of $8K$ samples for every methods, where $K$ is the number of classes. For TDA, the positive cache has size 5 while the negative cache is set to size 3 for each class.
\paragraph{Data augmentations.}
We note that our competitors use many computationally expensive augmentations in some settings. Since we do not propose to include such costly computations, we also do not use augmentations when running our competitors methods, so that we can compare performance at equal computational cost. Note that we also report the results of a non-online TTA method, MTA \cite{zanella_test-time_2024}, which relies on several augmentations of each image for informational purpose.
\paragraph{Prompts.} 
 First, we show results when applied on top of the zero-shot model with (i) handcrafted prompts (provided in Table \ref{tab:prompts} (Supplementary material)) and (ii) an ensemble of prompts (provided in Table \ref{tab:ct} (Supplementary material)). Then, we compare the methods when run  on top of few-shot adapted models with (i) prompt-tuning method CoOp \cite{zhou2022learning} and (ii) adapter method TaskRes \cite{yu2023task}. This comprehensive benchmarking highlights the broad applicability of OTTA methods and more specifically OGA across diverse scenarios. We aim to inspire other works to adopt a similar broad benchmarking methodology in future research.


\paragraph{Hyper-parameters.}
Our approach is dependent on a hyper-parameter $\nu$ (see Eq. \eqref{eq:map-prediction-rule-tau}). For the sake of generalization, we use the \textit{same fixed value} $\nu = 0.05$ across all datasets. We investigate its impact in Section \ref{sec:ablation}.

\paragraph{Evaluation metrics.} We report the average accuracy across 100 runs to mitigate variability in comparison due to the stochastic effects of data streams generation, which was not done in previous studies \cite{karmanov_efficient_2024, zhang_dual_2024} despite variability in results as demonstrated in Figure \ref{fig:abstract_figure} and Table \ref{tab:zero_shot_ACC}. Moreover, we argue that the latter metric is not sufficient to accurately compare methods and is not robust to \textit{tail risk}, where methods could show much worse accuracies for a small proportion of runs. The latter could make a method undesirable in practice. Therefore, we introduce a metric which we call \textit{Expected Tail Accuracy} (ETA) and is the average of accuracies in the $10\%$ worst cases, i.e.,
\begin{equation}
\text{ETA} = \dfrac{10}{N_{\text{runs}}} \sum_{r=1}^{N_\text{runs}}  \text{acc}^{(r)} \times \mathbbm{1} (\text{acc}^{(r)} \leq \text{acc}_{0.1})
\label{eq:expected-tail-accuracy}
\end{equation}
where $\text{acc}^{(r)}$ is the accuracy of run $r$ and $\text{acc}_{0.1}$ the lower $10^{\text{th}}$ percentile, and report this additional metric. Note that our approach does not contain any design choice for specifically mitigating these worst case accuracies and we just advocate for better performance reporting practices.

\begin{figure}[t]
\begin{center}
\includegraphics[width=0.5\textwidth]{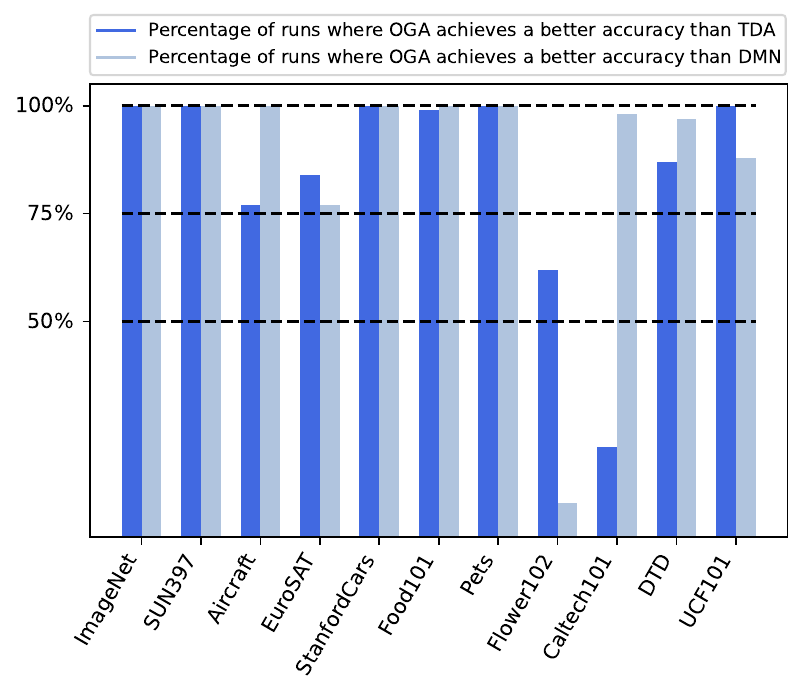}
\end{center}
\caption{For each dataset, we show the percentage of runs for which our method OGA achieves a higher accuracy than our competitors DMN and TDA. The experimental setting is the same as the one used for generating the results of Table \ref{tab:zero_shot_ACC_and_ETL}.}\label{fig:histograms_run_vs_run}
\end{figure}
\definecolor{CustomGreen}{RGB}{0,140, 0}
\definecolor{CustomRed}{RGB}{130,0,30}
\begin{table*}[t]
\centering
\caption{All methods are tested on the same 100 runs for each datasets with the same standard prompts of Table \ref{tab:prompts} (Supplementary Material). The data streams are i.i.d. and are processed in batches of 32 samples. The best metric is marked in \textbf{bold} while the second best is \underline{underlined}. For our method named OGA, we show the difference $\Delta \text{Competitor}$ with the best competitor.}
\label{tab:zero_shot_ACC_and_ETL}

\begin{subtable}{\linewidth}
\setlength\dashlinedash{0.2pt}
                        \setlength\dashlinegap{1.5pt}
                        \setlength\arrayrulewidth{0.3pt}
                        \renewcommand{\arraystretch}{1.32}
                            \caption{
 We report the average accuracy as well as the standard deviation over the 100 runs for each method and each dataset. As a reference, we provide the results of a non-online TTA method which relies on augmentations, namely MTA \cite{zanella_test-time_2024}.}
    \label{tab:zero_shot_ACC}
     \centering
    \resizebox{\textwidth}{!}{
        \begin{tabular}{l|ccccccccccc|c}
          \toprule
 & \rotatebox[origin=c]{45}{ImageNet} & \rotatebox[origin=c]{45}{SUN397} & \rotatebox[origin=c]{45}{Aircraft} & \rotatebox[origin=c]{45}{EuroSAT} & \rotatebox[origin=c]{45}{StanfordCars} & \rotatebox[origin=c]{45}{Food101} & \rotatebox[origin=c]{45}{Pets} & \rotatebox[origin=c]{45}{Flower102} & \rotatebox[origin=c]{45}{Caltech101} & \rotatebox[origin=c]{45}{DTD} & \rotatebox[origin=c]{45}{UCF101} & \rotatebox[origin=c]{0}{\textsc{Average}}
\\ \midrule
\textbf{Zero-Shot} & $\text{66.74}$ & $\text{62.55}$ & $\text{24.87}$ & $\text{48.25}$ & $\text{65.53}$ & $\text{85.88}$ & $\text{89.10}$ & $\text{70.81}$ & $\text{93.35}$ & $\text{43.32}$ & $\text{67.54}$ & $\text{65.3}$
\\ \midrule
MTA \scriptsize{(CVPR '24)} & 69.3 & 64.8 & 27.4 & 46.9 & 68.0 & 87.2 & 89.4 & 71.7 & 94.0 & 44.4 & 69.0 & 66.6
\\ \midrule
TDA \scriptsize{(CVPR '24)} & $\fontsize{11}{11}{\underline{\textcolor{Black}{67.9}}}$\fontsize{8}{8}{\thinspace$\pm 0.07$}&$\text{\textcolor{Black}{64.9}}$\fontsize{8}{8}{\thinspace$\pm 0.13$}&$\fontsize{11}{11}{\underline{\textcolor{Black}{24.7}}}$\fontsize{8}{8}{\thinspace$\pm 0.38$}&$\text{\textcolor{Black}{63.4}}$\fontsize{8}{8}{\thinspace$\pm 1.19$}&$\text{\textcolor{Black}{66.5}}$\fontsize{8}{8}{\thinspace$\pm 0.22$}&$\fontsize{11}{11}{\underline{\textcolor{Black}{85.8}}}$\fontsize{8}{8}{\thinspace$\pm 0.07$}&$\text{\textcolor{Black}{89.8}}$\fontsize{8}{8}{\thinspace$\pm 0.28$}&$\text{\textcolor{Black}{72.7}}$\fontsize{8}{8}{\thinspace$\pm 0.32$}&$\fontsize{11}{11}{\textbf{\textcolor{Black}{93.4}}}$\fontsize{8}{8}{\thinspace$\pm 0.36$}&$\fontsize{11}{11}{\underline{\textcolor{Black}{45.0}}}$\fontsize{8}{8}{\thinspace$\pm 0.41$}&$\text{\textcolor{Black}{70.5}}$\fontsize{8}{8}{\thinspace$\pm 0.35$}&$\underline{\text{\textcolor{Black}{67.7}}}$
\\ 
DMN \scriptsize{(CVPR '24)} &$\text{\textcolor{Black}{67.0}}$\fontsize{8}{8}{\thinspace$\pm 0.10$}&$\fontsize{11}{11}{\underline{\textcolor{Black}{64.9}}}$\fontsize{8}{8}{\thinspace$\pm 0.17$}&$\text{\textcolor{Black}{24.0}}$\fontsize{8}{8}{\thinspace$\pm 0.39$}&$\fontsize{11}{11}{\underline{\textcolor{Black}{64.0}}}$\fontsize{8}{8}{\thinspace$\pm 0.82$}&$\fontsize{11}{11}{\underline{\textcolor{Black}{67.0}}}$\fontsize{8}{8}{\thinspace$\pm 0.30$}&$\text{\textcolor{Black}{83.9}}$\fontsize{8}{8}{\thinspace$\pm 0.10$}&$\fontsize{11}{11}{\underline{\textcolor{Black}{89.9}}}$\fontsize{8}{8}{\thinspace$\pm 0.30$}&$\fontsize{11}{11}{\textbf{\textcolor{Black}{73.3}}}$\fontsize{8}{8}{\thinspace$\pm 0.34$}&$\text{\textcolor{Black}{92.6}}$\fontsize{8}{8}{\thinspace$\pm 0.42$}&$\text{\textcolor{Black}{44.7}}$\fontsize{8}{8}{\thinspace$\pm 0.62$}&$\fontsize{11}{11}{\underline{\textcolor{Black}{71.2}}}$\fontsize{8}{8}{\thinspace$\pm 0.41$}&$\text{\textcolor{Black}{67.5}}$
\\ 
\cellcolor{LightBlue}OGA (ours) &\cellcolor{LightBlue} $\fontsize{11}{11}{\textbf{\textcolor{Black}{68.5}}}$\fontsize{8}{8}{\thinspace$\pm 0.11$}&\cellcolor{LightBlue} $\fontsize{11}{11}{\textbf{\textcolor{Black}{66.0}}}$\fontsize{8}{8}{\thinspace$\pm 0.20$}&\cellcolor{LightBlue} $\fontsize{11}{11}{\textbf{\textcolor{Black}{25.3}}}$\fontsize{8}{8}{\thinspace$\pm 0.38$}&\cellcolor{LightBlue} $\fontsize{11}{11}{\textbf{\textcolor{Black}{64.5}}}$\fontsize{8}{8}{\thinspace$\pm 0.76$}&\cellcolor{LightBlue} $\fontsize{11}{11}{\textbf{\textcolor{Black}{67.8}}}$\fontsize{8}{8}{\thinspace$\pm 0.21$}&\cellcolor{LightBlue} $\fontsize{11}{11}{\textbf{\textcolor{Black}{86.1}}}$\fontsize{8}{8}{\thinspace$\pm 0.07$}&\cellcolor{LightBlue} $\fontsize{11}{11}{\textbf{\textcolor{Black}{91.7}}}$\fontsize{8}{8}{\thinspace$\pm 0.30$}&\cellcolor{LightBlue} $\fontsize{11}{11}{\underline{\textcolor{Black}{72.7}}}$\fontsize{8}{8}{\thinspace$\pm 0.38$}&\cellcolor{LightBlue} $\fontsize{11}{11}{\underline{\textcolor{Black}{93.2}}}$\fontsize{8}{8}{\thinspace$\pm 0.42$}&\cellcolor{LightBlue} $\fontsize{11}{11}{\textbf{\textcolor{Black}{45.8}}}$\fontsize{8}{8}{\thinspace$\pm 0.54$}&\cellcolor{LightBlue} $\fontsize{11}{11}{\textbf{\textcolor{Black}{71.6}}}$\fontsize{8}{8}{\thinspace$\pm 0.37$}&\cellcolor{LightBlue} $\fontsize{11}{11}{\textbf{\textcolor{Black}{68.5}}}$

\\\cellcolor{LightBlue}$\Delta$ \text{Competitor} &\cellcolor{LightBlue}  \textcolor{CustomGreen}{$+0.6$} &\cellcolor{LightBlue}  \textcolor{CustomGreen}{$+1.2$} &\cellcolor{LightBlue}  \textcolor{CustomGreen}{$+0.5$} &\cellcolor{LightBlue}  \textcolor{CustomGreen}{$+0.5$} &\cellcolor{LightBlue}  \textcolor{CustomGreen}{$+0.9$} &\cellcolor{LightBlue}  \textcolor{CustomGreen}{$+0.3$} &\cellcolor{LightBlue}  \textcolor{CustomGreen}{$+1.8$} &\cellcolor{LightBlue}  \textcolor{CustomRed}{$-0.6$} &\cellcolor{LightBlue}  \textcolor{CustomRed}{$-0.2$} &\cellcolor{LightBlue}  \textcolor{CustomGreen}{$+0.7$} &\cellcolor{LightBlue}  \textcolor{CustomGreen}{$+0.4$} &\cellcolor{LightBlue}  \textcolor{CustomGreen}{$+0.8$} 
\end{tabular}}

\end{subtable}

\vspace{0.6cm}

\begin{subtable}{\linewidth}
\setlength\dashlinedash{0.2pt}
                        \setlength\dashlinegap{1.5pt}
                        \setlength\arrayrulewidth{0.3pt}
                        \renewcommand{\arraystretch}{1.1}
                            \caption{
         We report the average accuracy over the 10 worst runs for each method and each dataset, i.e., the ETA (Equation \ref{eq:expected-tail-accuracy}). }
    \label{tab:zero_shot_ETL}
    \centering
     \resizebox{\textwidth}{!}{
        \begin{tabular}{l|ccccccccccc|c}
          \toprule
 & \rotatebox[origin=c]{45}{ImageNet} & \rotatebox[origin=c]{45}{SUN397} & \rotatebox[origin=c]{45}{Aircraft} & \rotatebox[origin=c]{45}{EuroSAT} & \rotatebox[origin=c]{45}{StanfordCars} & \rotatebox[origin=c]{45}{Food101} & \rotatebox[origin=c]{45}{Pets} & \rotatebox[origin=c]{45}{Flower102} & \rotatebox[origin=c]{45}{Caltech101} & \rotatebox[origin=c]{45}{DTD} & \rotatebox[origin=c]{45}{UCF101} & \rotatebox[origin=c]{0}{\textsc{Average}}
\\ \midrule
\textbf{Zero-Shot} & $\text{66.74}$ & $\text{62.55}$ & $\text{24.87}$ & $\text{48.25}$ & $\text{65.53}$ & $\text{85.88}$ & $\text{89.10}$ & $\text{70.81}$ & $\text{93.35}$ & $\text{43.32}$ & $\text{67.54}$ & $\text{65.3}$
\\ \midrule
TDA \scriptsize{(CVPR '24)} &${\underline{\textcolor{Black}{67.8}}}$&${\underline{\textcolor{Black}{64.6}}}$&${\underline{\textcolor{Black}{24.0}}}$&$\text{\textcolor{Black}{61.0}}$&$\text{\textcolor{Black}{66.1}}$&${\underline{\textcolor{Black}{85.7}}}$&${\underline{\textcolor{Black}{89.3}}}$&${\underline{\textcolor{Black}{72.1}}}$&${\textbf{\textcolor{Black}{92.5}}}$&${\underline{\textcolor{Black}{44.3}}}$&$\text{\textcolor{Black}{69.9}}$&$\underline{\text{\textcolor{Black}{67.0}}}$
\\ 
DMN \scriptsize{(CVPR '24)} &$\text{\textcolor{Black}{66.8}}$&$\text{\textcolor{Black}{64.6}}$&$\text{\textcolor{Black}{23.3}}$&${\underline{\textcolor{Black}{62.6}}}$&${\underline{\textcolor{Black}{66.4}}}$&$\text{\textcolor{Black}{83.7}}$&$\text{\textcolor{Black}{89.3}}$&${\textbf{\textcolor{Black}{72.7}}}$&$\text{\textcolor{Black}{91.6}}$&$\text{\textcolor{Black}{43.6}}$&${\underline{\textcolor{Black}{70.5}}}$&$\text{\textcolor{Black}{66.8}}$
\\ 
\cellcolor{LightBlue}OGA (ours) &\cellcolor{LightBlue}${\textbf{\textcolor{Black}{68.3}}}$&\cellcolor{LightBlue}${\textbf{\textcolor{Black}{65.7}}}$&\cellcolor{LightBlue}${\textbf{\textcolor{Black}{24.6}}}$&\cellcolor{LightBlue}${\textbf{\textcolor{Black}{63.2}}}$&\cellcolor{LightBlue}${\textbf{\textcolor{Black}{67.4}}}$&\cellcolor{LightBlue}${\textbf{\textcolor{Black}{85.9}}}$&\cellcolor{LightBlue}${\textbf{\textcolor{Black}{91.2}}}$&\cellcolor{LightBlue}$\text{\textcolor{Black}{71.9}}$&\cellcolor{LightBlue}${\underline{\textcolor{Black}{92.2}}}$&\cellcolor{LightBlue}${\textbf{\textcolor{Black}{44.9}}}$&\cellcolor{LightBlue}${\textbf{\textcolor{Black}{71.0}}}$&\cellcolor{LightBlue}$\textbf{\textcolor{Black}{67.9}}$

\end{tabular}}

\end{subtable}

\end{table*}

\section{Results and discussion}
\paragraph{Atop zero-shot.}
Table \ref{tab:zero_shot_ACC} shows that OGA performs better than OTTA competitors on 9 out of 11 datasets on average over 100 runs. For the two remaining datasets, our method still places second best. Note that each method is tested using the \textit{same} 100 runs for each dataset, and that we use the same \textit{fixed hyper-parameters} for all datasets. Overall, this proves the effectiveness of our approach. Now we analyse the results to the light of our proposed metric ETA. Notice in Tables \ref{tab:zero_shot_ACC} and \ref{tab:zero_shot_ETL} that on several datasets (ImageNet, SUN397, StanfordCars, Pets), the ETA of our method is higher than the average accuracy of our competitors, i.e. the worst $10\%$ runs for our method still ranks higher than the average of our competitors.

Moreover, Table \ref{tab:zero_shot_ACC} shows the ETA of all methods are lower than the zero-shot performance of CLIP on the Aircraft dataset, indicating that they quite often deliver performance below zero-shot. This breakdown demonstrates the value of ETA in providing deeper insights into the results. We also report the accuracy of a non-online state-of-the-art TTA method, MTA \cite{zanella_boosting_2024}, which relies on multiple augmentations of the input images and does not retain information from samples. This shows how casting the problem of TTA in an online setting can be highly beneficial, with a striking example being the dramatic gain of more than 15 points of accuracy on EuroSAT.
Meanwhile, Figure \ref{fig:histograms_run_vs_run} shows the percentage of runs for which OGA achieves a higher accuracy than TDA and DMN for each dataset. Observe that for 5 datasets (ImageNet, SUN397, StanfordCars, Pets and UCF101), our approach achieves a higher accuracy than TDA for all of the 100 runs used for testing. In comparison with DMN, our method yields a higher accuracy for all of the runs for 6 datasets (ImageNet, SUN397, Aircraft, StanfordCars, Food101, Pets). Finally, we compare the three methods in the same setting but with the ensemble of prompts of Table \ref{tab:ct} (Supplementary Material) in the Table \ref{tab:zero_shot_ensemble_ACC}. In this experiment, our method ranks first for 8 datasets out of 11, and second on the remaining three. Therefore, our approach is robust to changes in the prompts used for zero-shot predictions, a finding further confirmed in the next paragraph.
\setlength\dashlinedash{0.2pt}
                        \setlength\dashlinegap{1.5pt}
                        \setlength\arrayrulewidth{0.3pt}
                        \renewcommand{\arraystretch}{1.2}
                        \begin{table*}[t]
                            \caption{Reported performance is the averaged accuracy over the same 100 runs for each method and each dataset. We use the custom prompts ensemble (see Table \ref{tab:ct} of the Supplementary Material).}
    \label{tab:zero_shot_ensemble_ACC}
    \centering
       \resizebox{\textwidth}{!}{
        \begin{tabular}{l|ccccccccccc|c}
          \toprule
 & \rotatebox[origin=c]{45}{ImageNet} & \rotatebox[origin=c]{45}{SUN397} & \rotatebox[origin=c]{45}{Aircraft} & \rotatebox[origin=c]{45}{EuroSAT} & \rotatebox[origin=c]{45}{StanfordCars} & \rotatebox[origin=c]{45}{Food101} & \rotatebox[origin=c]{45}{Pets} & \rotatebox[origin=c]{45}{Flower102} & \rotatebox[origin=c]{45}{Caltech101} & \rotatebox[origin=c]{45}{DTD} & \rotatebox[origin=c]{45}{UCF101} & \rotatebox[origin=c]{0}{\textsc{Average}}
\\ \midrule
\textbf{Zero-Shot} & $\text{68.73}$ & $\text{66.17}$ & $\text{23.10}$ & $\text{50.54}$ & $\text{66.05}$ & $\text{85.59}$ & $\text{87.90}$ & $\text{67.07}$ & $\text{93.87}$ & $\text{45.15}$ & $\text{67.59}$ & $\text{65.6}$
\\ \midrule
TDA \scriptsize{(CVPR '24)} &$\fontsize{11}{11}{\underline{\textcolor{Black}{69.3}}}$\fontsize{8}{8}{\thinspace$\pm 0.06$}&$\fontsize{11}{11}{\underline{\textcolor{Black}{67.5}}}$\fontsize{8}{8}{\thinspace$\pm 0.11$}&$\fontsize{11}{11}{\underline{\textcolor{Black}{23.1}}}$\fontsize{8}{8}{\thinspace$\pm 0.30$}&$\fontsize{11}{11}{\textbf{\textcolor{Black}{57.3}}}$\fontsize{8}{8}{\thinspace$\pm 0.68$}&$\text{\textcolor{Black}{66.8}}$\fontsize{8}{8}{\thinspace$\pm 0.23$}&$\fontsize{11}{11}{\underline{\textcolor{Black}{85.3}}}$\fontsize{8}{8}{\thinspace$\pm 0.06$}&$\fontsize{11}{11}{\underline{\textcolor{Black}{87.9}}}$\fontsize{8}{8}{\thinspace$\pm 0.22$}&$\text{\textcolor{Black}{68.7}}$\fontsize{8}{8}{\thinspace$\pm 0.39$}&$\fontsize{11}{11}{\textbf{\textcolor{Black}{94.0}}}$\fontsize{8}{8}{\thinspace$\pm 0.33$}&$\text{\textcolor{Black}{46.4}}$\fontsize{8}{8}{\thinspace$\pm 0.39$}&$\text{\textcolor{Black}{69.7}}$\fontsize{8}{8}{\thinspace$\pm 0.33$}&$\underline{\text{\textcolor{Black}{66.9}}}$
\\ 
DMN \scriptsize{(CVPR '24)} &$\text{\textcolor{Black}{68.2}}$\fontsize{8}{8}{\thinspace$\pm 0.10$}&$\text{\textcolor{Black}{66.9}}$\fontsize{8}{8}{\thinspace$\pm 0.16$}&$\text{\textcolor{Black}{22.7}}$\fontsize{8}{8}{\thinspace$\pm 0.34$}&$\text{\textcolor{Black}{51.7}}$\fontsize{8}{8}{\thinspace$\pm 1.25$}&$\fontsize{11}{11}{\underline{\textcolor{Black}{67.5}}}$\fontsize{8}{8}{\thinspace$\pm 0.28$}&$\text{\textcolor{Black}{83.5}}$\fontsize{8}{8}{\thinspace$\pm 0.09$}&$\text{\textcolor{Black}{87.8}}$\fontsize{8}{8}{\thinspace$\pm 0.33$}&$\fontsize{11}{11}{\textbf{\textcolor{Black}{71.0}}}$\fontsize{8}{8}{\thinspace$\pm 0.43$}&$\text{\textcolor{Black}{93.2}}$\fontsize{8}{8}{\thinspace$\pm 0.46$}&$\fontsize{11}{11}{\underline{\textcolor{Black}{47.0}}}$\fontsize{8}{8}{\thinspace$\pm 0.61$}&$\fontsize{11}{11}{\underline{\textcolor{Black}{70.6}}}$\fontsize{8}{8}{\thinspace$\pm 0.43$}&$\text{\textcolor{Black}{66.4}}$
\\ 
\cellcolor{LightBlue}OGA (ours) &\cellcolor{LightBlue} $\fontsize{11}{11}{\textbf{\textcolor{Black}{69.4}}}$\fontsize{8}{8}{\thinspace$\pm 0.11$}&\cellcolor{LightBlue} $\fontsize{11}{11}{\textbf{\textcolor{Black}{67.9}}}$\fontsize{8}{8}{\thinspace$\pm 0.16$}&\cellcolor{LightBlue} $\fontsize{11}{11}{\textbf{\textcolor{Black}{23.2}}}$\fontsize{8}{8}{\thinspace$\pm 0.39$}&\cellcolor{LightBlue} $\fontsize{11}{11}{\underline{\textcolor{Black}{54.2}}}$\fontsize{8}{8}{\thinspace$\pm 1.38$}&\cellcolor{LightBlue} $\fontsize{11}{11}{\textbf{\textcolor{Black}{68.1}}}$\fontsize{8}{8}{\thinspace$\pm 0.20$}&\cellcolor{LightBlue} $\fontsize{11}{11}{\textbf{\textcolor{Black}{85.6}}}$\fontsize{8}{8}{\thinspace$\pm 0.07$}&\cellcolor{LightBlue} $\fontsize{11}{11}{\textbf{\textcolor{Black}{89.4}}}$\fontsize{8}{8}{\thinspace$\pm 0.26$}&\cellcolor{LightBlue} $\fontsize{11}{11}{\underline{\textcolor{Black}{69.2}}}$\fontsize{8}{8}{\thinspace$\pm 0.40$}&\cellcolor{LightBlue} $\fontsize{11}{11}{\underline{\textcolor{Black}{93.6}}}$\fontsize{8}{8}{\thinspace$\pm 0.40$}&\cellcolor{LightBlue} $\fontsize{11}{11}{\textbf{\textcolor{Black}{47.9}}}$\fontsize{8}{8}{\thinspace$\pm 0.44$}&\cellcolor{LightBlue} $\fontsize{11}{11}{\textbf{\textcolor{Black}{71.4}}}$\fontsize{8}{8}{\thinspace$\pm 0.41$}&\cellcolor{LightBlue} $\textbf{\textcolor{Black}{67.3}}$

\end{tabular}}

\end{table*}

\aboverulesep = 0.1mm 
\belowrulesep = 0.1mm 

\begin{table*}
\centering
\caption{Methods are tested on top of popular few-shot methods. For each few-shot method, we train three adapted models using three different random seeds and run OTTA methods on the same 100 runs per seed. The data streams are i.i.d. and are processed in batches of 32 samples. Reported performance is the averaged accuracy and standard deviation over the resulting 300 runs. The best metric is marked in \textbf{bold} while the second best is \underline{underlined}.}


\label{tab:atop}

\begin{subtable}{\linewidth}
\setlength\dashlinedash{0.2pt}
                        \setlength\dashlinegap{1.5pt}
                        \setlength\arrayrulewidth{0.3pt}
                        \renewcommand{\arraystretch}{1.2}
                             \caption{CoOp \cite{zhou2022learning} is a popular prompt-tuning method for few-shots adaptation, which adds learnable tokens to to the texts defining the classes (see Equation \ref{eq:coop_and_taskres}).}
 
    \label{tab:atop_coop}
    \centering
    \resizebox{\textwidth}{!}{
        \begin{tabular}{ll|ccccccccccc}
          \toprule
 &  & \rotatebox[origin=c]{45}{ImageNet} & \rotatebox[origin=c]{45}{SUN397} & \rotatebox[origin=c]{45}{Aircraft} & \rotatebox[origin=c]{45}{EuroSAT} & \rotatebox[origin=c]{45}{StanfordCars} & \rotatebox[origin=c]{45}{Food101} & \rotatebox[origin=c]{45}{Pets} & \rotatebox[origin=c]{45}{Flower102} & \rotatebox[origin=c]{45}{Caltech101} & \rotatebox[origin=c]{45}{DTD} & \rotatebox[origin=c]{45}{UCF101} 
\\ \midrule
\multirow{4}{*}{\rotatebox[origin=c]{90}{1 shot}} &
CoOp &$\text{\textcolor{Black}{65.7}}$&$\text{\textcolor{Black}{66.9}}$&$\text{\textcolor{Black}{20.8}}$&$\text{\textcolor{Black}{56.4}}$&$\text{\textcolor{Black}{67.5}}$&$\text{\textcolor{Black}{84.3}}$&$\text{\textcolor{Black}{90.2}}$&$\text{\textcolor{Black}{78.3}}$&$\text{\textcolor{Black}{92.5}}$&$\text{\textcolor{Black}{50.1}}$&$\text{\textcolor{Black}{71.2}}$
\\ 
& 
$^+$ TDA \scriptsize{(CVPR '24)} &$\fontsize{11}{11}{\underline{\textcolor{Black}{66.8}}}$\fontsize{8}{8}{\thinspace$\pm 0.04$}&$\fontsize{11}{11}{\underline{\textcolor{Black}{68.2}}}$\fontsize{8}{8}{\thinspace$\pm 0.07$}&$\fontsize{11}{11}{\underline{\textcolor{Black}{21.9}}}$\fontsize{8}{8}{\thinspace$\pm 0.24$}&$\fontsize{11}{11}{\underline{\textcolor{Black}{61.7}}}$\fontsize{8}{8}{\thinspace$\pm 0.38$}&$\text{\textcolor{Black}{68.1}}$\fontsize{8}{8}{\thinspace$\pm 0.15$}&$\fontsize{11}{11}{\underline{\textcolor{Black}{84.6}}}$\fontsize{8}{8}{\thinspace$\pm 0.05$}&$\fontsize{11}{11}{\underline{\textcolor{Black}{90.4}}}$\fontsize{8}{8}{\thinspace$\pm 0.14$}&$\text{\textcolor{Black}{80.7}}$\fontsize{8}{8}{\thinspace$\pm 0.23$}&$\fontsize{11}{11}{\textbf{\textcolor{Black}{92.9}}}$\fontsize{8}{8}{\thinspace$\pm 0.29$}&$\text{\textcolor{Black}{51.5}}$\fontsize{8}{8}{\thinspace$\pm 0.21$}&$\text{\textcolor{Black}{73.1}}$\fontsize{8}{8}{\thinspace$\pm 0.17$}
\\ 
& 
$^+$ DMN \scriptsize{(CVPR '24)} &$\text{\textcolor{Black}{66.5}}$\fontsize{8}{8}{\thinspace$\pm 0.07$}&$\text{\textcolor{Black}{68.2}}$\fontsize{8}{8}{\thinspace$\pm 0.11$}&$\text{\textcolor{Black}{21.6}}$\fontsize{8}{8}{\thinspace$\pm 0.24$}&$\fontsize{11}{11}{\textbf{\textcolor{Black}{62.0}}}$\fontsize{8}{8}{\thinspace$\pm 0.55$}&$\fontsize{11}{11}{\underline{\textcolor{Black}{69.0}}}$\fontsize{8}{8}{\thinspace$\pm 0.21$}&$\text{\textcolor{Black}{83.7}}$\fontsize{8}{8}{\thinspace$\pm 0.07$}&$\text{\textcolor{Black}{89.9}}$\fontsize{8}{8}{\thinspace$\pm 0.20$}&$\fontsize{11}{11}{\textbf{\textcolor{Black}{82.7}}}$\fontsize{8}{8}{\thinspace$\pm 0.24$}&$\text{\textcolor{Black}{92.6}}$\fontsize{8}{8}{\thinspace$\pm 0.34$}&$\fontsize{11}{11}{\underline{\textcolor{Black}{52.0}}}$\fontsize{8}{8}{\thinspace$\pm 0.39$}&$\fontsize{11}{11}{\underline{\textcolor{Black}{73.8}}}$\fontsize{8}{8}{\thinspace$\pm 0.34$}
\\ 
& 
\cellcolor{LightBlue}$^+$ OGA (ours) &\cellcolor{LightBlue} $\fontsize{11}{11}{\textbf{\textcolor{Black}{67.6}}}$\fontsize{8}{8}{\thinspace$\pm 0.07$}&\cellcolor{LightBlue} $\fontsize{11}{11}{\textbf{\textcolor{Black}{69.1}}}$\fontsize{8}{8}{\thinspace$\pm 0.12$}&\cellcolor{LightBlue} $\fontsize{11}{11}{\textbf{\textcolor{Black}{22.1}}}$\fontsize{8}{8}{\thinspace$\pm 0.20$}&\cellcolor{LightBlue} $\text{\textcolor{Black}{61.6}}$\fontsize{8}{8}{\thinspace$\pm 0.59$}&\cellcolor{LightBlue} $\fontsize{11}{11}{\textbf{\textcolor{Black}{69.6}}}$\fontsize{8}{8}{\thinspace$\pm 0.13$}&\cellcolor{LightBlue} $\fontsize{11}{11}{\textbf{\textcolor{Black}{85.2}}}$\fontsize{8}{8}{\thinspace$\pm 0.05$}&\cellcolor{LightBlue} $\fontsize{11}{11}{\textbf{\textcolor{Black}{91.3}}}$\fontsize{8}{8}{\thinspace$\pm 0.17$}&\cellcolor{LightBlue} $\fontsize{11}{11}{\underline{\textcolor{Black}{81.0}}}$\fontsize{8}{8}{\thinspace$\pm 0.30$}&\cellcolor{LightBlue} $\fontsize{11}{11}{\underline{\textcolor{Black}{92.8}}}$\fontsize{8}{8}{\thinspace$\pm 0.33$}&\cellcolor{LightBlue} $\fontsize{11}{11}{\textbf{\textcolor{Black}{52.5}}}$\fontsize{8}{8}{\thinspace$\pm 0.34$}&\cellcolor{LightBlue} $\fontsize{11}{11}{\textbf{\textcolor{Black}{74.3}}}$\fontsize{8}{8}{\thinspace$\pm 0.25$}
\\
\cmidrule[0.5pt](lr{0em}){1-13}
\multirow{4}{*}{\rotatebox[origin=c]{90}{4 shots}} &
CoOp &$\text{\textcolor{Black}{68.8}}$&$\text{\textcolor{Black}{69.7}}$&$\text{\textcolor{Black}{30.8}}$&$\text{\textcolor{Black}{69.7}}$&$\text{\textcolor{Black}{74.4}}$&$\text{\textcolor{Black}{84.3}}$&$\fontsize{11}{11}{\underline{\textcolor{Black}{92.5}}}$&$\text{\textcolor{Black}{92.2}}$&$\text{\textcolor{Black}{94.5}}$&$\text{\textcolor{Black}{59.4}}$&$\text{\textcolor{Black}{77.5}}$
\\ 
& 
$^+$ TDA \scriptsize{(CVPR '24)} &$\fontsize{11}{11}{\underline{\textcolor{Black}{69.4}}}$\fontsize{8}{8}{\thinspace$\pm 0.04$}&$\fontsize{11}{11}{\underline{\textcolor{Black}{70.6}}}$\fontsize{8}{8}{\thinspace$\pm 0.06$}&$\fontsize{11}{11}{\underline{\textcolor{Black}{31.2}}}$\fontsize{8}{8}{\thinspace$\pm 0.20$}&$\text{\textcolor{Black}{73.7}}$\fontsize{8}{8}{\thinspace$\pm 0.34$}&$\fontsize{11}{11}{\underline{\textcolor{Black}{74.8}}}$\fontsize{8}{8}{\thinspace$\pm 0.13$}&$\fontsize{11}{11}{\underline{\textcolor{Black}{84.9}}}$\fontsize{8}{8}{\thinspace$\pm 0.03$}&$\text{\textcolor{Black}{92.4}}$\fontsize{8}{8}{\thinspace$\pm 0.12$}&$\text{\textcolor{Black}{92.9}}$\fontsize{8}{8}{\thinspace$\pm 0.14$}&$\fontsize{11}{11}{\underline{\textcolor{Black}{94.5}}}$\fontsize{8}{8}{\thinspace$\pm 0.30$}&$\text{\textcolor{Black}{60.9}}$\fontsize{8}{8}{\thinspace$\pm 0.19$}&$\text{\textcolor{Black}{78.9}}$\fontsize{8}{8}{\thinspace$\pm 0.17$}
\\ 
& 
$^+$ DMN \scriptsize{(CVPR '24)} &$\text{\textcolor{Black}{68.6}}$\fontsize{8}{8}{\thinspace$\pm 0.06$}&$\text{\textcolor{Black}{70.5}}$\fontsize{8}{8}{\thinspace$\pm 0.09$}&$\text{\textcolor{Black}{31.0}}$\fontsize{8}{8}{\thinspace$\pm 0.25$}&$\fontsize{11}{11}{\underline{\textcolor{Black}{73.8}}}$\fontsize{8}{8}{\thinspace$\pm 0.40$}&$\text{\textcolor{Black}{74.6}}$\fontsize{8}{8}{\thinspace$\pm 0.18$}&$\text{\textcolor{Black}{84.0}}$\fontsize{8}{8}{\thinspace$\pm 0.07$}&$\text{\textcolor{Black}{91.7}}$\fontsize{8}{8}{\thinspace$\pm 0.17$}&$\fontsize{11}{11}{\textbf{\textcolor{Black}{93.4}}}$\fontsize{8}{8}{\thinspace$\pm 0.14$}&$\text{\textcolor{Black}{94.4}}$\fontsize{8}{8}{\thinspace$\pm 0.31$}&$\fontsize{11}{11}{\underline{\textcolor{Black}{61.3}}}$\fontsize{8}{8}{\thinspace$\pm 0.28$}&$\fontsize{11}{11}{\underline{\textcolor{Black}{79.1}}}$\fontsize{8}{8}{\thinspace$\pm 0.28$}
\\ 
& 
\cellcolor{LightBlue}$^+$ OGA (ours) &\cellcolor{LightBlue} $\fontsize{11}{11}{\textbf{\textcolor{Black}{69.7}}}$\fontsize{8}{8}{\thinspace$\pm 0.06$}&\cellcolor{LightBlue} $\fontsize{11}{11}{\textbf{\textcolor{Black}{71.5}}}$\fontsize{8}{8}{\thinspace$\pm 0.11$}&\cellcolor{LightBlue} $\fontsize{11}{11}{\textbf{\textcolor{Black}{31.7}}}$\fontsize{8}{8}{\thinspace$\pm 0.24$}&\cellcolor{LightBlue} $\fontsize{11}{11}{\textbf{\textcolor{Black}{75.3}}}$\fontsize{8}{8}{\thinspace$\pm 0.38$}&\cellcolor{LightBlue} $\fontsize{11}{11}{\textbf{\textcolor{Black}{76.1}}}$\fontsize{8}{8}{\thinspace$\pm 0.12$}&\cellcolor{LightBlue} $\fontsize{11}{11}{\textbf{\textcolor{Black}{84.9}}}$\fontsize{8}{8}{\thinspace$\pm 0.07$}&\cellcolor{LightBlue} $\fontsize{11}{11}{\textbf{\textcolor{Black}{93.0}}}$\fontsize{8}{8}{\thinspace$\pm 0.14$}&\cellcolor{LightBlue} $\fontsize{11}{11}{\underline{\textcolor{Black}{92.9}}}$\fontsize{8}{8}{\thinspace$\pm 0.23$}&\cellcolor{LightBlue} $\fontsize{11}{11}{\textbf{\textcolor{Black}{94.5}}}$\fontsize{8}{8}{\thinspace$\pm 0.31$}&\cellcolor{LightBlue} $\fontsize{11}{11}{\textbf{\textcolor{Black}{61.6}}}$\fontsize{8}{8}{\thinspace$\pm 0.27$}&\cellcolor{LightBlue} $\fontsize{11}{11}{\textbf{\textcolor{Black}{79.8}}}$\fontsize{8}{8}{\thinspace$\pm 0.21$}
\\
\end{tabular}}

\end{subtable}

\vspace{0.6cm}

\begin{subtable}{\linewidth}

\setlength\dashlinedash{0.2pt}
                        \setlength\dashlinegap{1.5pt}
                        \setlength\arrayrulewidth{0.3pt}
                        \renewcommand{\arraystretch}{1.2}
                             \caption{
         TaskRes \cite{yu2023task} is a popular adapter method which adds a bias to the text embedding of each class (see Equation \ref{eq:coop_and_taskres}). }
    \label{tab:atop_taskres}
    \centering
    \resizebox{\textwidth}{!}{
               \begin{tabular}{ll|ccccccccccc}
          \toprule
 &  & \rotatebox[origin=c]{45}{ImageNet} & \rotatebox[origin=c]{45}{SUN397} & \rotatebox[origin=c]{45}{Aircraft} & \rotatebox[origin=c]{45}{EuroSAT} & \rotatebox[origin=c]{45}{StanfordCars} & \rotatebox[origin=c]{45}{Food101} & \rotatebox[origin=c]{45}{Pets} & \rotatebox[origin=c]{45}{Flower102} & \rotatebox[origin=c]{45}{Caltech101} & \rotatebox[origin=c]{45}{DTD} & \rotatebox[origin=c]{45}{UCF101} 
\\ \midrule
\multirow{4}{*}{\rotatebox[origin=c]{90}{1 shot}} &
TaskRes &$\text{\textcolor{Black}{69.6}}$&$\text{\textcolor{Black}{68.1}}$&$\fontsize{11}{11}{\underline{\textcolor{Black}{31.2}}}$&$\text{\textcolor{Black}{65.7}}$&$\fontsize{11}{11}{\underline{\textcolor{Black}{69.1}}}$&$\text{\textcolor{Black}{84.5}}$&$\fontsize{11}{11}{\underline{\textcolor{Black}{90.1}}}$&$\text{\textcolor{Black}{81.6}}$&$\fontsize{11}{11}{\underline{\textcolor{Black}{93.6}}}$&$\text{\textcolor{Black}{53.4}}$&$\text{\textcolor{Black}{71.8}}$
\\ 
& 
$^+$ TDA \scriptsize{(CVPR '24)} &$\fontsize{11}{11}{\textbf{\textcolor{Black}{70.1}}}$\fontsize{8}{8}{\thinspace$\pm 0.06$}&$\fontsize{11}{11}{\underline{\textcolor{Black}{69.3}}}$\fontsize{8}{8}{\thinspace$\pm 0.08$}&$\text{\textcolor{Black}{30.7}}$\fontsize{8}{8}{\thinspace$\pm 0.22$}&$\fontsize{11}{11}{\underline{\textcolor{Black}{69.5}}}$\fontsize{8}{8}{\thinspace$\pm 0.44$}&$\text{\textcolor{Black}{68.9}}$\fontsize{8}{8}{\thinspace$\pm 0.15$}&$\fontsize{11}{11}{\underline{\textcolor{Black}{84.9}}}$\fontsize{8}{8}{\thinspace$\pm 0.04$}&$\text{\textcolor{Black}{90.1}}$\fontsize{8}{8}{\thinspace$\pm 0.15$}&$\text{\textcolor{Black}{83.6}}$\fontsize{8}{8}{\thinspace$\pm 0.30$}&$\fontsize{11}{11}{\textbf{\textcolor{Black}{93.9}}}$\fontsize{8}{8}{\thinspace$\pm 0.27$}&$\fontsize{11}{11}{\underline{\textcolor{Black}{55.3}}}$\fontsize{8}{8}{\thinspace$\pm 0.27$}&$\fontsize{11}{11}{\underline{\textcolor{Black}{73.1}}}$\fontsize{8}{8}{\thinspace$\pm 0.20$}
\\ 
& 
$^+$ DMN \scriptsize{(CVPR '24)} &$\text{\textcolor{Black}{68.7}}$\fontsize{8}{8}{\thinspace$\pm 0.09$}&$\text{\textcolor{Black}{68.2}}$\fontsize{8}{8}{\thinspace$\pm 0.12$}&$\text{\textcolor{Black}{30.2}}$\fontsize{8}{8}{\thinspace$\pm 0.26$}&$\text{\textcolor{Black}{69.5}}$\fontsize{8}{8}{\thinspace$\pm 0.54$}&$\text{\textcolor{Black}{68.9}}$\fontsize{8}{8}{\thinspace$\pm 0.18$}&$\text{\textcolor{Black}{83.4}}$\fontsize{8}{8}{\thinspace$\pm 0.09$}&$\text{\textcolor{Black}{89.4}}$\fontsize{8}{8}{\thinspace$\pm 0.19$}&$\fontsize{11}{11}{\textbf{\textcolor{Black}{85.9}}}$\fontsize{8}{8}{\thinspace$\pm 0.29$}&$\text{\textcolor{Black}{92.8}}$\fontsize{8}{8}{\thinspace$\pm 0.38$}&$\text{\textcolor{Black}{54.4}}$\fontsize{8}{8}{\thinspace$\pm 0.44$}&$\text{\textcolor{Black}{72.8}}$\fontsize{8}{8}{\thinspace$\pm 0.30$}
\\ 
& 
\cellcolor{LightBlue}$^+$ OGA (ours) &\cellcolor{LightBlue} $\fontsize{11}{11}{\underline{\textcolor{Black}{69.9}}}$\fontsize{8}{8}{\thinspace$\pm 0.09$}&\cellcolor{LightBlue} $\fontsize{11}{11}{\textbf{\textcolor{Black}{69.4}}}$\fontsize{8}{8}{\thinspace$\pm 0.14$}&\cellcolor{LightBlue} $\fontsize{11}{11}{\textbf{\textcolor{Black}{31.5}}}$\fontsize{8}{8}{\thinspace$\pm 0.24$}&\cellcolor{LightBlue} $\fontsize{11}{11}{\textbf{\textcolor{Black}{70.6}}}$\fontsize{8}{8}{\thinspace$\pm 0.53$}&\cellcolor{LightBlue} $\fontsize{11}{11}{\textbf{\textcolor{Black}{70.9}}}$\fontsize{8}{8}{\thinspace$\pm 0.12$}&\cellcolor{LightBlue} $\fontsize{11}{11}{\textbf{\textcolor{Black}{85.5}}}$\fontsize{8}{8}{\thinspace$\pm 0.09$}&\cellcolor{LightBlue} $\fontsize{11}{11}{\textbf{\textcolor{Black}{91.3}}}$\fontsize{8}{8}{\thinspace$\pm 0.18$}&\cellcolor{LightBlue} $\fontsize{11}{11}{\underline{\textcolor{Black}{84.1}}}$\fontsize{8}{8}{\thinspace$\pm 0.33$}&\cellcolor{LightBlue} $\text{\textcolor{Black}{93.4}}$\fontsize{8}{8}{\thinspace$\pm 0.35$}&\cellcolor{LightBlue} $\fontsize{11}{11}{\textbf{\textcolor{Black}{55.8}}}$\fontsize{8}{8}{\thinspace$\pm 0.40$}&\cellcolor{LightBlue} $\fontsize{11}{11}{\textbf{\textcolor{Black}{73.9}}}$\fontsize{8}{8}{\thinspace$\pm 0.26$}
\\
\cmidrule[0.5pt](lr{0em}){1-13}
\multirow{4}{*}{\rotatebox[origin=c]{90}{4 shots}} &
TaskRes &$\fontsize{11}{11}{\underline{\textcolor{Black}{71.0}}}$&$\fontsize{11}{11}{\underline{\textcolor{Black}{72.8}}}$&$\fontsize{11}{11}{\underline{\textcolor{Black}{33.2}}}$&$\text{\textcolor{Black}{73.9}}$&$\fontsize{11}{11}{\underline{\textcolor{Black}{76.1}}}$&$\fontsize{11}{11}{\underline{\textcolor{Black}{86.1}}}$&$\fontsize{11}{11}{\underline{\textcolor{Black}{91.9}}}$&$\text{\textcolor{Black}{85.0}}$&$\fontsize{11}{11}{\underline{\textcolor{Black}{94.8}}}$&$\text{\textcolor{Black}{59.6}}$&$\text{\textcolor{Black}{75.5}}$
\\ 
& 
$^+$ TDA \scriptsize{(CVPR '24)} &$\fontsize{11}{11}{\textbf{\textcolor{Black}{71.3}}}$\fontsize{8}{8}{\thinspace$\pm 0.05$}&$\fontsize{11}{11}{\textbf{\textcolor{Black}{73.2}}}$\fontsize{8}{8}{\thinspace$\pm 0.06$}&$\text{\textcolor{Black}{32.9}}$\fontsize{8}{8}{\thinspace$\pm 0.24$}&$\fontsize{11}{11}{\textbf{\textcolor{Black}{76.1}}}$\fontsize{8}{8}{\thinspace$\pm 0.31$}&$\text{\textcolor{Black}{75.3}}$\fontsize{8}{8}{\thinspace$\pm 0.15$}&$\text{\textcolor{Black}{85.9}}$\fontsize{8}{8}{\thinspace$\pm 0.04$}&$\text{\textcolor{Black}{91.6}}$\fontsize{8}{8}{\thinspace$\pm 0.14$}&$\text{\textcolor{Black}{87.3}}$\fontsize{8}{8}{\thinspace$\pm 0.29$}&$\fontsize{11}{11}{\textbf{\textcolor{Black}{94.9}}}$\fontsize{8}{8}{\thinspace$\pm 0.31$}&$\fontsize{11}{11}{\underline{\textcolor{Black}{61.3}}}$\fontsize{8}{8}{\thinspace$\pm 0.30$}&$\fontsize{11}{11}{\underline{\textcolor{Black}{76.6}}}$\fontsize{8}{8}{\thinspace$\pm 0.20$}
\\ 
& 
$^+$ DMN \scriptsize{(CVPR '24)} &$\text{\textcolor{Black}{69.6}}$\fontsize{8}{8}{\thinspace$\pm 0.08$}&$\text{\textcolor{Black}{71.5}}$\fontsize{8}{8}{\thinspace$\pm 0.11$}&$\text{\textcolor{Black}{32.1}}$\fontsize{8}{8}{\thinspace$\pm 0.26$}&$\text{\textcolor{Black}{73.5}}$\fontsize{8}{8}{\thinspace$\pm 0.43$}&$\text{\textcolor{Black}{74.5}}$\fontsize{8}{8}{\thinspace$\pm 0.19$}&$\text{\textcolor{Black}{83.9}}$\fontsize{8}{8}{\thinspace$\pm 0.09$}&$\text{\textcolor{Black}{90.6}}$\fontsize{8}{8}{\thinspace$\pm 0.18$}&$\fontsize{11}{11}{\textbf{\textcolor{Black}{88.7}}}$\fontsize{8}{8}{\thinspace$\pm 0.28$}&$\text{\textcolor{Black}{94.2}}$\fontsize{8}{8}{\thinspace$\pm 0.36$}&$\text{\textcolor{Black}{59.5}}$\fontsize{8}{8}{\thinspace$\pm 0.41$}&$\text{\textcolor{Black}{75.9}}$\fontsize{8}{8}{\thinspace$\pm 0.30$}
\\ 
& 
\cellcolor{LightBlue}$^+$ OGA (ours) &\cellcolor{LightBlue} $\text{\textcolor{Black}{70.7}}$\fontsize{8}{8}{\thinspace$\pm 0.09$}&\cellcolor{LightBlue} $\text{\textcolor{Black}{72.6}}$\fontsize{8}{8}{\thinspace$\pm 0.13$}&\cellcolor{LightBlue} $\fontsize{11}{11}{\textbf{\textcolor{Black}{33.5}}}$\fontsize{8}{8}{\thinspace$\pm 0.26$}&\cellcolor{LightBlue} $\fontsize{11}{11}{\underline{\textcolor{Black}{74.4}}}$\fontsize{8}{8}{\thinspace$\pm 0.49$}&\cellcolor{LightBlue} $\fontsize{11}{11}{\textbf{\textcolor{Black}{77.4}}}$\fontsize{8}{8}{\thinspace$\pm 0.12$}&\cellcolor{LightBlue} $\fontsize{11}{11}{\textbf{\textcolor{Black}{86.2}}}$\fontsize{8}{8}{\thinspace$\pm 0.07$}&\cellcolor{LightBlue} $\fontsize{11}{11}{\textbf{\textcolor{Black}{92.3}}}$\fontsize{8}{8}{\thinspace$\pm 0.18$}&\cellcolor{LightBlue} $\fontsize{11}{11}{\underline{\textcolor{Black}{87.3}}}$\fontsize{8}{8}{\thinspace$\pm 0.36$}&\cellcolor{LightBlue} $\text{\textcolor{Black}{94.7}}$\fontsize{8}{8}{\thinspace$\pm 0.30$}&\cellcolor{LightBlue} $\fontsize{11}{11}{\textbf{\textcolor{Black}{61.8}}}$\fontsize{8}{8}{\thinspace$\pm 0.37$}&\cellcolor{LightBlue} $\fontsize{11}{11}{\textbf{\textcolor{Black}{77.2}}}$\fontsize{8}{8}{\thinspace$\pm 0.26$}
\\
\end{tabular}}

\end{subtable}

\end{table*}
\paragraph{Atop few-shot.}
In Table \ref{tab:atop}, we report the results atop two popular few-shot adaptation methods. For CoOp (Table \ref{tab:atop_coop}), a prompt-learning method, our approach yields the strongest improvement, performing better on average for 8 datasets out of 11 in the 1-shot setting and for 10 out of 11 datasets in the 4-shot setting. For TaskRes (Table \ref{tab:atop_taskres}), an adapter method, our approach also achieves the highest overall accuracy gain, ranking first for 8 datasets out of 11 in the 1-shot setting. In the 4-shot setting, our method achieves highest accuracy for 6 datasets out of 11. Interestingly, we observe that the few-shot adaptation reduces the variability of OTTA method on nearly every dataset. Finally, we see that in the vast majority of the cases, the OTTA methods improve over the few-shot adapted model, which proves the benefits of using OTTA atop adapted models.


\section{Ablation studies}
\label{sec:ablation}
\paragraph{Likelihood weighting hyper-parameter $\nu$.}
\begin{figure}[t]
\begin{center}
\includegraphics[width=0.5\textwidth]{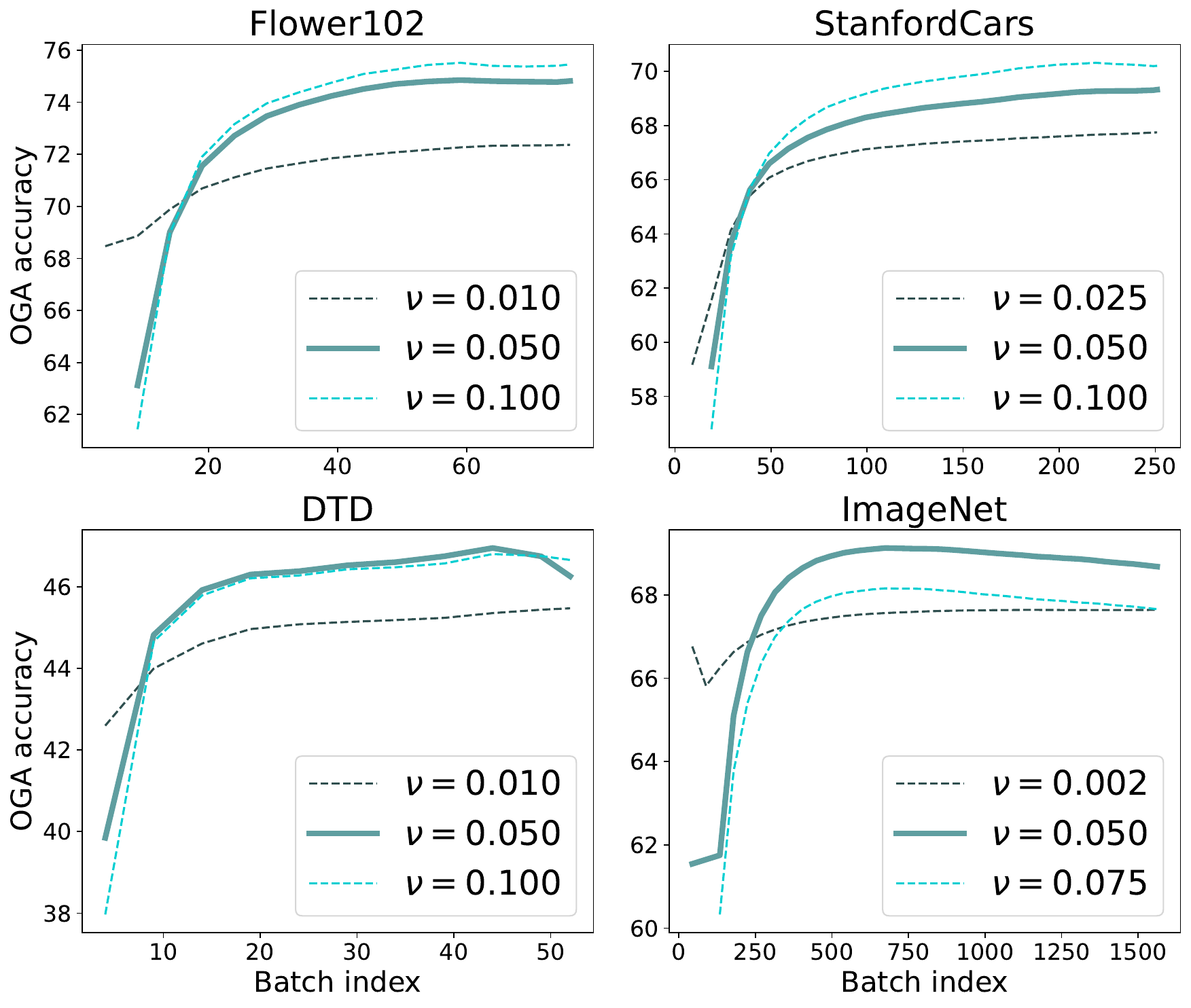}
\end{center}
\caption{We show the dynamic of the accuracy of our OGA method as it starts from an empty cache, averaged on 100 runs. At regular intervals, we evaluate the accuracy of OGA on the complete test set. }\label{fig:ablation_tau}
\end{figure}
Our method uses the same fixed hyper-parameter $\nu = 0.05$ (see Equation \ref{eq:map-prediction-rule-tau}) for all experiments and datasets. It controls the degree to which the Gaussian likelihood is pushed away from the uniform distribution. Therefore, when $\nu = 0$, our MAP degenerates to the zero-shot prior. Following, it is expected that higher values of $\nu$ are detrimental when the Gaussian modeling is poor (e.g., at the beginning of a run). Figure \ref{fig:ablation_tau} illustrates that our choice of hyper-parameter is essentially a trade-off between mitigating early transitory effects, when the cache is either empty or filled with poor quality samples, and end point accuracy. This interesting observation could pave the way for improving our method by designing an adaptive rule for $\nu$ dependent on the state of the cache.
\paragraph{Size of the cache.}
We show results with different cache sizes in Table \ref{tab:ablation_n}, i.e., the maximum number of cached samples per class. This illustrates how the cache size is a trade-off between diversity and contamination with incorrectly labeled samples.

\begin{table}[!h] 
\small
    \caption{Ablation study on the size of the cache for our method. We report the averaged accuracy over the 11 datasets.}
    \label{tab:ablation_n}
    \centering
        \begin{tabular}{l|c}
            \toprule
            & \rotatebox[origin=c]{0}{\textsc{Average}} \\
            \midrule
            \textbf{Zero-Shot} & $\text{65.3}$ \\
            \midrule
            OGA w/ cache size 4 & $\text{\textcolor{Black}{67.8}}$ \\
            \cellcolor{LightBlue}OGA w/ cache size 8 & \cellcolor{LightBlue} $\textbf{\textcolor{Black}{68.5}}$ \\
            OGA w/ cache size 16 & $\text{\textcolor{Black}{68.0}}$ \\
            OGA w/ cache size 32 & $\text{\textcolor{Black}{67.0}}$ \\
        \end{tabular}
    
\end{table}
\paragraph{Precision matrix estimation.}
We show that it is beneficial to use different estimators depending on the number of samples in the cache. To do so, we run our method either with only the Ridge estimator or only the (pseudo-)inverse, and present results in Table~\ref{tab:ablation_P}.


\begin{table}[t] 
\small
    \caption{Ablation study on the use of two different estimators instead of one. We report the averaged accuracy over the 11 datasets.}
    \label{tab:ablation_P}
    \centering
        \begin{tabular}{l|c}
            \toprule
 & \rotatebox[origin=c]{0}{\textsc{Average}}
\\ \midrule
\textbf{Zero-Shot} & $\text{65.3}$
\\ \midrule
\cellcolor{LightBlue}OGA w/ Ridge and Inverse &\cellcolor{LightBlue}$\textbf{\textcolor{Black}{68.5}}$
\\ 
OGA w/ only (pseudo-)inverse &$\text{\textcolor{Black}{66.6}}$
\\ 
OGA w/ only Ridge &$\text{\textcolor{Black}{68.3}}$
        \end{tabular}
    
\end{table}

\paragraph{Batch size.}
In all experiments, we process the data streams in batches of 32 samples. In Table~\ref{tab:ablation_bs}, we show that our method is able to process the streams sample by sample and that it benefits from increased batch sizes. The latter is due to the fact that cache-based methods are quicker to fill their cache with quality samples when the batch size increases, as the cache is updated before predicting. Note our approach still achieves a higher average accuracy in batch size 1 compared to our competitors in batch size 32.

\begin{table}[t] 
\small
    \caption{Ablation study on the batch size used to process the data streams. We report the averaged accuracy over the 11 datasets. }
    \label{tab:ablation_bs}
    \centering
        \begin{tabular}{l|c}
            \toprule
            & \rotatebox[origin=c]{0}{\textsc{Average}} \\
            \midrule
            \textbf{Zero-Shot} & $\text{65.3}$ \\
            \midrule
            OGA w/ batch size 1 & $\text{\textcolor{Black}{68.42}}$ \\
            \cellcolor{LightBlue} OGA w/ batch size 32 & \cellcolor{LightBlue} $\text{\textcolor{Black}{68.46}}$ \\
            OGA w/ batch size 64 & $\text{\textcolor{Black}{68.53}}$ \\
            OGA w/ batch size 128 & $\text{\textcolor{Black}{68.61}}$ \\
        \end{tabular}
    
\end{table}


\section{Conclusion}
In this study, we proposed Online Gaussian Adaptation (OGA), a method for the online test-time-adaptation of VLMs. Our method uses a modeling of the class-conditional likelihoods of visual features with multivariate Gaussians, which are estimated from low-entropy samples collected along the data stream. We compared our approach to state-of-the-art methods with a rigorous evaluation protocol, inspired by the significant variability in the measured accuracy observed between runs. Using 100 runs per dataset and our proposed \textit{Expected Tail Accuracy} (ETA) metric which captures the performance in worst-case scenarios, we showed that our method delivers strong performance with \textit{fixed hyper-parameters} across datasets. Lastly, we showed that applying OTTA methods on top of few-shot learning methods, either prompt-tuning or adapter, is highly beneficial. We hope our work will encourage more rigorous and diverse evaluation practices in the OTTA community. 
\paragraph{Future works.}
As highlighted in our ablation study, an interesting avenue to explore seem to be the design of an adaptive rule for our hyper-parameter $\nu$ (Equation \ref{eq:map-prediction-rule-tau}), depending on the state of the cache, as well as the strength of the zero-shot prior or of the few-shot adaptation.

\section{Acknowledgments}
 C.~Fuchs is funded by the MedReSyst
project, supported by FEDER and the Walloon Region. M.~Zanella is funded by the Walloon region under grant No.~2010235 (ARIAC by DIGITALWALLONIA4.AI). Part of the computational resources have been provided by the Consortium des Équipements de Calcul Intensif (CÉCI), funded by the Fonds de la Recherche Scientifique de Belgique (F.R.S.-FNRS) under Grant No. 2.5020.11 and by the Walloon Region.

{
    \small
    \bibliographystyle{ieeenat_fullname}
    \bibliography{main}
}

\clearpage
\setcounter{page}{1}
\onecolumn

\appendix

\begin{center}
    {\Large \textbf{Online Gaussian Test-Time Adaptation of Vision-Language Models}}\\[10pt]
    {\Large Supplementary Material}\\[5pt]
\end{center}

\section{Prompts}
We show the handcrafted prompts used in the relevant experiments in Table \ref{tab:app_all_prompts}.
\begin{table}[h]
\caption{Prompt templates for each dataset.} 
\label{tab:app_all_prompts}
\centering
\begin{subtable}{.5\textwidth}

\centering

\subcaption{Prompt templates used in the experiments unless otherwise specified.}\label{tab:prompts}
\resizebox{0.9\linewidth}{!}{
 \begin{tabular}{ccc}
    \toprule
    Dataset     &  Prompt template \\
    \midrule
   ImageNet     &   "\texttt{a photo of a [].}" \\
   SUN397     &   "\texttt{a photo of a [].}" \\
    Aircraft    &  "\texttt{a photo of a [], a type of aircraft.}",\\
    EuroSAT    & "\texttt{a centered satellite photo of [].}", \\
    Cars    &  "\texttt{a photo of a [].}", \\
    Food101    &  "\texttt{a photo of [], a type of food.}",\\
    Pets    &  "\texttt{a photo of [], a type of pet.}", \\
    Flower102    &  "\texttt{a photo of a [], a type of flower.}",\\
    Caltech101    &  "\texttt{a photo of a [].}", \\
    DTD    &  "\texttt{[] texture.}", \\
    UCF101    &  "\texttt{a photo of a person doing [].}",\\
    \bottomrule
    \end{tabular}}
\end{subtable}
\hfill
\begin{subtable}{.4\textwidth}
\centering
\subcaption{Custom prompt templates ensemble.}\label{tab:ct} 

\centering
\resizebox{0.8\linewidth}{!}{
   \begin{tabular}{c}
    \toprule
   "\texttt{itap of a [].}" \\
   "\texttt{a bad photo of the [].}" \\
   "\texttt{a origami [].}" \\
   "\texttt{a photo of the large [].}" \\
   "\texttt{a [] in a video game.}" \\
   "\texttt{art of the [].}" \\
   "\texttt{a photo of the small [].}" \\
    \bottomrule
    \end{tabular}}

\end{subtable}

\end{table}
\vspace{0.5cm}

\section{Results with different architectures.}
In the main paper, all experiments are done using the ViT-B/16 version of CLIP. Here, we show that results with other backbones (ViT-L/14, ViT-B/32, ResNet50 and ResNet101), presented in Tables \ref{tab:vits}, \ref{tab:vits_ensemble}, \ref{tab:resnets}, \ref{tab:resnets_ensemble} and \ref{tab:summary_backbones}, are coherent with the observations made previously. Note that we use the same \textit{fixed hyper-parameters} across all datasets and architectures. For each dataset, the methods are tested using the same 100 runs.

\subsection{Results with other ViT architectures.}
\begin{table*}[h]
\centering
\caption{We show results obtained with other ViT-based architectures and the prompts of Table \ref{tab:prompts}. }

\label{tab:vits}

\begin{subtable}{\linewidth}

\setlength\dashlinedash{0.2pt}
                        \setlength\dashlinegap{1.5pt}
                        \setlength\arrayrulewidth{0.3pt}
                        \renewcommand{\arraystretch}{1.2}
                            \caption{
        With ViT-B/32.}
    \label{tab:vit_b32}
    \centering
    \resizebox{\textwidth}{!}{
        \begin{tabular}{l|cccccccccccc}
          \toprule
 & \rotatebox[origin=c]{45}{ImageNet} & \rotatebox[origin=c]{45}{SUN397} & \rotatebox[origin=c]{45}{Aircraft} & \rotatebox[origin=c]{45}{EuroSAT} & \rotatebox[origin=c]{45}{StanfordCars} & \rotatebox[origin=c]{45}{Food101} & \rotatebox[origin=c]{45}{Pets} & \rotatebox[origin=c]{45}{Flower102} & \rotatebox[origin=c]{45}{Caltech101} & \rotatebox[origin=c]{45}{DTD} & \rotatebox[origin=c]{45}{UCF101} & \rotatebox[origin=c]{45}{Average}
\\ \midrule
\textbf{Zero-Shot} & $\text{62.03}$ & $\text{62.11}$ & $\text{19.14}$ & $\text{45.38}$ & $\text{60.17}$ & $\text{80.40}$ & $\text{87.33}$ & $\text{66.67}$ & $\text{91.44}$ & $\text{42.61}$ & $\text{63.52}$ & $\text{61.9}$
\\ \midrule
TDA \scriptsize{(CVPR '24)} &$\fontsize{11}{11}{\underline{\textcolor{Black}{62.8}}}$\fontsize{8}{8}{\thinspace$\pm 0.07$}&$\fontsize{11}{11}{\underline{\textcolor{Black}{63.7}}}$\fontsize{8}{8}{\thinspace$\pm 0.12$}&$\text{\textcolor{Black}{18.4}}$\fontsize{8}{8}{\thinspace$\pm 0.33$}&$\text{\textcolor{Black}{46.3}}$\fontsize{8}{8}{\thinspace$\pm 0.93$}&$\fontsize{11}{11}{\underline{\textcolor{Black}{60.3}}}$\fontsize{8}{8}{\thinspace$\pm 0.26$}&$\fontsize{11}{11}{\underline{\textcolor{Black}{80.0}}}$\fontsize{8}{8}{\thinspace$\pm 0.06$}&$\fontsize{11}{11}{\underline{\textcolor{Black}{86.7}}}$\fontsize{8}{8}{\thinspace$\pm 0.27$}&$\text{\textcolor{Black}{67.6}}$\fontsize{8}{8}{\thinspace$\pm 0.31$}&$\fontsize{11}{11}{\textbf{\textcolor{Black}{91.0}}}$\fontsize{8}{8}{\thinspace$\pm 0.41$}&$\fontsize{11}{11}{\underline{\textcolor{Black}{43.2}}}$\fontsize{8}{8}{\thinspace$\pm 0.46$}&$\fontsize{11}{11}{\textbf{\textcolor{Black}{65.3}}}$\fontsize{8}{8}{\thinspace$\pm 0.32$}&$\underline{\textcolor{Black}{62.3}}$
\\ 
DMN \scriptsize{(CVPR '24)} &$\text{\textcolor{Black}{61.5}}$\fontsize{8}{8}{\thinspace$\pm 0.12$}&$\text{\textcolor{Black}{63.4}}$\fontsize{8}{8}{\thinspace$\pm 0.19$}&$\fontsize{11}{11}{\underline{\textcolor{Black}{18.4}}}$\fontsize{8}{8}{\thinspace$\pm 0.29$}&$\fontsize{11}{11}{\underline{\textcolor{Black}{47.5}}}$\fontsize{8}{8}{\thinspace$\pm 1.17$}&$\text{\textcolor{Black}{60.0}}$\fontsize{8}{8}{\thinspace$\pm 0.31$}&$\text{\textcolor{Black}{77.5}}$\fontsize{8}{8}{\thinspace$\pm 0.12$}&$\text{\textcolor{Black}{86.6}}$\fontsize{8}{8}{\thinspace$\pm 0.35$}&$\fontsize{11}{11}{\textbf{\textcolor{Black}{68.1}}}$\fontsize{8}{8}{\thinspace$\pm 0.32$}&$\text{\textcolor{Black}{89.2}}$\fontsize{8}{8}{\thinspace$\pm 0.63$}&$\text{\textcolor{Black}{42.9}}$\fontsize{8}{8}{\thinspace$\pm 0.68$}&$\text{\textcolor{Black}{64.8}}$\fontsize{8}{8}{\thinspace$\pm 0.49$}&$\text{\textcolor{Black}{61.8}}$
\\ 
\cellcolor{LightBlue} OGA (ours) &\cellcolor{LightBlue} $\fontsize{11}{11}{\textbf{\textcolor{Black}{63.0}}}$\fontsize{8}{8}{\thinspace$\pm 0.11$}&\cellcolor{LightBlue} $\fontsize{11}{11}{\textbf{\textcolor{Black}{64.5}}}$\fontsize{8}{8}{\thinspace$\pm 0.16$}&\cellcolor{LightBlue} $\fontsize{11}{11}{\textbf{\textcolor{Black}{18.7}}}$\fontsize{8}{8}{\thinspace$\pm 0.32$}&\cellcolor{LightBlue} $\fontsize{11}{11}{\textbf{\textcolor{Black}{49.3}}}$\fontsize{8}{8}{\thinspace$\pm 1.09$}&\cellcolor{LightBlue} $\fontsize{11}{11}{\textbf{\textcolor{Black}{61.6}}}$\fontsize{8}{8}{\thinspace$\pm 0.20$}&\cellcolor{LightBlue} $\fontsize{11}{11}{\textbf{\textcolor{Black}{80.1}}}$\fontsize{8}{8}{\thinspace$\pm 0.10$}&\cellcolor{LightBlue} $\fontsize{11}{11}{\textbf{\textcolor{Black}{88.2}}}$\fontsize{8}{8}{\thinspace$\pm 0.29$}&\cellcolor{LightBlue} $\fontsize{11}{11}{\underline{\textcolor{Black}{67.8}}}$\fontsize{8}{8}{\thinspace$\pm 0.31$}&\cellcolor{LightBlue} $\fontsize{11}{11}{\underline{\textcolor{Black}{89.5}}}$\fontsize{8}{8}{\thinspace$\pm 0.56$}&\cellcolor{LightBlue} $\fontsize{11}{11}{\textbf{\textcolor{Black}{44.2}}}$\fontsize{8}{8}{\thinspace$\pm 0.54$}&\cellcolor{LightBlue} $\fontsize{11}{11}{\underline{\textcolor{Black}{65.2}}}$\fontsize{8}{8}{\thinspace$\pm 0.37$}&\cellcolor{LightBlue} $\textbf{\textcolor{Black}{62.9}}$
\end{tabular}}

\end{subtable}

\vspace{0.3cm}

\begin{subtable}{\linewidth}
\setlength\dashlinedash{0.2pt}
                        \setlength\dashlinegap{1.5pt}
                        \setlength\arrayrulewidth{0.3pt}
                        \renewcommand{\arraystretch}{1.2}
 \caption{
        With ViT-L/14.}
    \label{tab:vit_l14}
    \centering
    \resizebox{\textwidth}{!}{
        \begin{tabular}{l|cccccccccccc}
          \toprule
 & \rotatebox[origin=c]{45}{ImageNet} & \rotatebox[origin=c]{45}{SUN397} & \rotatebox[origin=c]{45}{Aircraft} & \rotatebox[origin=c]{45}{EuroSAT} & \rotatebox[origin=c]{45}{StanfordCars} & \rotatebox[origin=c]{45}{Food101} & \rotatebox[origin=c]{45}{Pets} & \rotatebox[origin=c]{45}{Flower102} & \rotatebox[origin=c]{45}{Caltech101} & \rotatebox[origin=c]{45}{DTD} & \rotatebox[origin=c]{45}{UCF101} & \rotatebox[origin=c]{45}{Average}
\\ \midrule
\textbf{Zero-Shot} & $\text{73.44}$ & $\text{67.66}$ & $\text{32.52}$ & $\text{60.27}$ & $\text{76.89}$ & $\text{90.92}$ & $\text{93.49}$ & $\text{79.58}$ & $\text{95.21}$ & $\text{53.43}$ & $\text{75.05}$ & $\text{72.6}$
\\ \midrule
TDA \scriptsize{(CVPR '24)} &$\fontsize{11}{11}{\underline{\textcolor{Black}{74.4}}}$\fontsize{8}{8}{\thinspace$\pm 0.05$}&$\text{\textcolor{Black}{69.3}}$\fontsize{8}{8}{\thinspace$\pm 0.10$}&$\fontsize{11}{11}{\underline{\textcolor{Black}{32.8}}}$\fontsize{8}{8}{\thinspace$\pm 0.41$}&$\text{\textcolor{Black}{63.9}}$\fontsize{8}{8}{\thinspace$\pm 0.81$}&$\text{\textcolor{Black}{77.0}}$\fontsize{8}{8}{\thinspace$\pm 0.25$}&$\fontsize{11}{11}{\textbf{\textcolor{Black}{90.8}}}$\fontsize{8}{8}{\thinspace$\pm 0.05$}&$\fontsize{11}{11}{\underline{\textcolor{Black}{93.5}}}$\fontsize{8}{8}{\thinspace$\pm 0.15$}&$\text{\textcolor{Black}{80.3}}$\fontsize{8}{8}{\thinspace$\pm 0.32$}&$\fontsize{11}{11}{\underline{\textcolor{Black}{94.5}}}$\fontsize{8}{8}{\thinspace$\pm 0.33$}&$\fontsize{11}{11}{\underline{\textcolor{Black}{55.0}}}$\fontsize{8}{8}{\thinspace$\pm 0.37$}&$\text{\textcolor{Black}{76.7}}$\fontsize{8}{8}{\thinspace$\pm 0.27$}&${\textcolor{Black}{73.5}}$
\\ 
DMN \scriptsize{(CVPR '24)} &$\text{\textcolor{Black}{74.4}}$\fontsize{8}{8}{\thinspace$\pm 0.10$}&$\fontsize{11}{11}{\underline{\textcolor{Black}{70.0}}}$\fontsize{8}{8}{\thinspace$\pm 0.16$}&$\text{\textcolor{Black}{32.3}}$\fontsize{8}{8}{\thinspace$\pm 0.46$}&$\fontsize{11}{11}{\textbf{\textcolor{Black}{64.1}}}$\fontsize{8}{8}{\thinspace$\pm 0.79$}&$\fontsize{11}{11}{\underline{\textcolor{Black}{78.1}}}$\fontsize{8}{8}{\thinspace$\pm 0.31$}&$\text{\textcolor{Black}{89.8}}$\fontsize{8}{8}{\thinspace$\pm 0.09$}&$\text{\textcolor{Black}{93.1}}$\fontsize{8}{8}{\thinspace$\pm 0.23$}&$\fontsize{11}{11}{\textbf{\textcolor{Black}{81.6}}}$\fontsize{8}{8}{\thinspace$\pm 0.31$}&$\text{\textcolor{Black}{94.4}}$\fontsize{8}{8}{\thinspace$\pm 0.43$}&$\text{\textcolor{Black}{54.5}}$\fontsize{8}{8}{\thinspace$\pm 0.62$}&$\fontsize{11}{11}{\underline{\textcolor{Black}{78.1}}}$\fontsize{8}{8}{\thinspace$\pm 0.40$}&$\underline{\textcolor{Black}{73.7}}$
\\ 
\cellcolor{LightBlue} OGA (ours) &\cellcolor{LightBlue} $\fontsize{11}{11}{\textbf{\textcolor{Black}{75.2}}}$\fontsize{8}{8}{\thinspace$\pm 0.12$}&\cellcolor{LightBlue} $\fontsize{11}{11}{\textbf{\textcolor{Black}{70.7}}}$\fontsize{8}{8}{\thinspace$\pm 0.19$}&\cellcolor{LightBlue} $\fontsize{11}{11}{\textbf{\textcolor{Black}{33.2}}}$\fontsize{8}{8}{\thinspace$\pm 0.57$}&\cellcolor{LightBlue} $\fontsize{11}{11}{\underline{\textcolor{Black}{63.9}}}$\fontsize{8}{8}{\thinspace$\pm 0.93$}&\cellcolor{LightBlue} $\fontsize{11}{11}{\textbf{\textcolor{Black}{79.2}}}$\fontsize{8}{8}{\thinspace$\pm 0.29$}&\cellcolor{LightBlue} $\fontsize{11}{11}{\underline{\textcolor{Black}{90.7}}}$\fontsize{8}{8}{\thinspace$\pm 0.08$}&\cellcolor{LightBlue} $\fontsize{11}{11}{\textbf{\textcolor{Black}{93.9}}}$\fontsize{8}{8}{\thinspace$\pm 0.18$}&\cellcolor{LightBlue} $\fontsize{11}{11}{\underline{\textcolor{Black}{81.3}}}$\fontsize{8}{8}{\thinspace$\pm 0.34$}&\cellcolor{LightBlue} $\fontsize{11}{11}{\textbf{\textcolor{Black}{94.9}}}$\fontsize{8}{8}{\thinspace$\pm 0.37$}&\cellcolor{LightBlue} $\fontsize{11}{11}{\textbf{\textcolor{Black}{56.1}}}$\fontsize{8}{8}{\thinspace$\pm 0.60$}&\cellcolor{LightBlue} $\fontsize{11}{11}{\textbf{\textcolor{Black}{78.4}}}$\fontsize{8}{8}{\thinspace$\pm 0.36$}&\cellcolor{LightBlue} $\textbf{\textcolor{Black}{74.3}}$
\end{tabular}}

\end{subtable}

\end{table*}
\begin{table*}[!ht]
\centering
\caption{We show results obtained with other ViT-based architectures and the custom ensemble of prompts of Table \ref{tab:ct}. }

\label{tab:vits_ensemble}

\begin{subtable}{\linewidth}

\setlength\dashlinedash{0.2pt}
                        \setlength\dashlinegap{1.5pt}
                        \setlength\arrayrulewidth{0.3pt}
                        \renewcommand{\arraystretch}{1.2}
                            \caption{
        With ViT-B/32.}
    \label{tab:vitB32_ensemble}
    \centering
    \resizebox{\textwidth}{!}{
        \begin{tabular}{l|cccccccccccc}
          \toprule
 & \rotatebox[origin=c]{45}{ImageNet} & \rotatebox[origin=c]{45}{SUN397} & \rotatebox[origin=c]{45}{Aircraft} & \rotatebox[origin=c]{45}{EuroSAT} & \rotatebox[origin=c]{45}{StanfordCars} & \rotatebox[origin=c]{45}{Food101} & \rotatebox[origin=c]{45}{Pets} & \rotatebox[origin=c]{45}{Flower102} & \rotatebox[origin=c]{45}{Caltech101} & \rotatebox[origin=c]{45}{DTD} & \rotatebox[origin=c]{45}{UCF101} & \rotatebox[origin=c]{45}{Average}
\\ \midrule
\textbf{Zero-Shot} & $\text{63.74}$ & $\text{63.99}$ & $\text{18.39}$ & $\text{43.00}$ & $\text{60.14}$ & $\text{79.78}$ & $\text{84.96}$ & $\text{63.62}$ & $\text{92.17}$ & $\text{43.20}$ & $\text{62.09}$ & $\text{61.4}$
\\ \midrule
TDA \scriptsize{(CVPR '24)} &$\fontsize{11}{11}{\textbf{\textcolor{Black}{64.1}}}$\fontsize{8}{8}{\thinspace$\pm 0.07$}&$\fontsize{11}{11}{\underline{\textcolor{Black}{65.3}}}$\fontsize{8}{8}{\thinspace$\pm 0.11$}&$\fontsize{11}{11}{\underline{\textcolor{Black}{17.6}}}$\fontsize{8}{8}{\thinspace$\pm 0.35$}&$\fontsize{11}{11}{\underline{\textcolor{Black}{49.5}}}$\fontsize{8}{8}{\thinspace$\pm 1.15$}&$\fontsize{11}{11}{\underline{\textcolor{Black}{60.6}}}$\fontsize{8}{8}{\thinspace$\pm 0.26$}&$\fontsize{11}{11}{\underline{\textcolor{Black}{79.4}}}$\fontsize{8}{8}{\thinspace$\pm 0.08$}&$\fontsize{11}{11}{\underline{\textcolor{Black}{84.0}}}$\fontsize{8}{8}{\thinspace$\pm 0.29$}&$\text{\textcolor{Black}{64.1}}$\fontsize{8}{8}{\thinspace$\pm 0.29$}&$\fontsize{11}{11}{\textbf{\textcolor{Black}{91.8}}}$\fontsize{8}{8}{\thinspace$\pm 0.34$}&$\text{\textcolor{Black}{44.8}}$\fontsize{8}{8}{\thinspace$\pm 0.48$}&$\text{\textcolor{Black}{64.2}}$\fontsize{8}{8}{\thinspace$\pm 0.28$}&${\textcolor{Black}{62.3}}$
\\ 
DMN \scriptsize{(CVPR '24)} &$\text{\textcolor{Black}{62.4}}$\fontsize{8}{8}{\thinspace$\pm 0.12$}&$\text{\textcolor{Black}{64.6}}$\fontsize{8}{8}{\thinspace$\pm 0.18$}&$\text{\textcolor{Black}{17.4}}$\fontsize{8}{8}{\thinspace$\pm 0.35$}&$\text{\textcolor{Black}{46.2}}$\fontsize{8}{8}{\thinspace$\pm 1.41$}&$\text{\textcolor{Black}{60.3}}$\fontsize{8}{8}{\thinspace$\pm 0.33$}&$\text{\textcolor{Black}{77.0}}$\fontsize{8}{8}{\thinspace$\pm 0.11$}&$\text{\textcolor{Black}{83.7}}$\fontsize{8}{8}{\thinspace$\pm 0.35$}&$\fontsize{11}{11}{\textbf{\textcolor{Black}{65.4}}}$\fontsize{8}{8}{\thinspace$\pm 0.35$}&$\text{\textcolor{Black}{90.0}}$\fontsize{8}{8}{\thinspace$\pm 0.55$}&$\fontsize{11}{11}{\underline{\textcolor{Black}{45.6}}}$\fontsize{8}{8}{\thinspace$\pm 0.60$}&$\fontsize{11}{11}{\underline{\textcolor{Black}{65.5}}}$\fontsize{8}{8}{\thinspace$\pm 0.42$}&${\textcolor{Black}{61.6}}$
\\ 
\cellcolor{LightBlue}OGA (ours) &\cellcolor{LightBlue} $\fontsize{11}{11}{\underline{\textcolor{Black}{63.7}}}$\fontsize{8}{8}{\thinspace$\pm 0.10$}&\cellcolor{LightBlue} $\fontsize{11}{11}{\textbf{\textcolor{Black}{65.4}}}$\fontsize{8}{8}{\thinspace$\pm 0.18$}&\cellcolor{LightBlue} $\fontsize{11}{11}{\textbf{\textcolor{Black}{18.3}}}$\fontsize{8}{8}{\thinspace$\pm 0.31$}&\cellcolor{LightBlue} $\fontsize{11}{11}{\textbf{\textcolor{Black}{49.5}}}$\fontsize{8}{8}{\thinspace$\pm 1.15$}&\cellcolor{LightBlue} $\fontsize{11}{11}{\textbf{\textcolor{Black}{61.5}}}$\fontsize{8}{8}{\thinspace$\pm 0.23$}&\cellcolor{LightBlue} $\fontsize{11}{11}{\textbf{\textcolor{Black}{79.4}}}$\fontsize{8}{8}{\thinspace$\pm 0.10$}&\cellcolor{LightBlue} $\fontsize{11}{11}{\textbf{\textcolor{Black}{85.9}}}$\fontsize{8}{8}{\thinspace$\pm 0.27$}&\cellcolor{LightBlue} $\fontsize{11}{11}{\underline{\textcolor{Black}{64.4}}}$\fontsize{8}{8}{\thinspace$\pm 0.42$}&\cellcolor{LightBlue} $\fontsize{11}{11}{\underline{\textcolor{Black}{90.2}}}$\fontsize{8}{8}{\thinspace$\pm 0.56$}&\cellcolor{LightBlue} $\fontsize{11}{11}{\textbf{\textcolor{Black}{46.5}}}$\fontsize{8}{8}{\thinspace$\pm 0.58$}&\cellcolor{LightBlue} $\fontsize{11}{11}{\textbf{\textcolor{Black}{65.6}}}$\fontsize{8}{8}{\thinspace$\pm 0.34$}&\cellcolor{LightBlue} $\textbf{\textcolor{Black}{62.8}}$ 
\end{tabular}}

\end{subtable}

\vspace{0.1cm}

\begin{subtable}{\linewidth}
 
\setlength\dashlinedash{0.2pt}
                        \setlength\dashlinegap{1.5pt}
                        \setlength\arrayrulewidth{0.3pt}
                        \renewcommand{\arraystretch}{1.2}
                            \caption{
        With ViT-L/14.}
    \label{tab:vitl14_ensemble}
    \centering
    \resizebox{\textwidth}{!}{
        \begin{tabular}{l|cccccccccccc}
          \toprule
 & \rotatebox[origin=c]{45}{ImageNet} & \rotatebox[origin=c]{45}{SUN397} & \rotatebox[origin=c]{45}{Aircraft} & \rotatebox[origin=c]{45}{EuroSAT} & \rotatebox[origin=c]{45}{StanfordCars} & \rotatebox[origin=c]{45}{Food101} & \rotatebox[origin=c]{45}{Pets} & \rotatebox[origin=c]{45}{Flower102} & \rotatebox[origin=c]{45}{Caltech101} & \rotatebox[origin=c]{45}{DTD} & \rotatebox[origin=c]{45}{UCF101} & \rotatebox[origin=c]{45}{Average}
\\ \midrule
\textbf{Zero-Shot} & $\text{75.90}$ & $\text{70.44}$ & $\text{31.23}$ & $\text{50.16}$ & $\text{77.68}$ & $\text{91.29}$ & $\text{92.83}$ & $\text{77.47}$ & $\text{95.58}$ & $\text{55.73}$ & $\text{76.29}$ & $\text{72.2}$
\\ \midrule
TDA \scriptsize{(CVPR '24)} &$\fontsize{11}{11}{\underline{\textcolor{Black}{76.3}}}$\fontsize{8}{8}{\thinspace$\pm 0.05$}&$\fontsize{11}{11}{\underline{\textcolor{Black}{71.5}}}$\fontsize{8}{8}{\thinspace$\pm 0.11$}&$\text{\textcolor{Black}{31.3}}$\fontsize{8}{8}{\thinspace$\pm 0.40$}&$\text{\textcolor{Black}{63.5}}$\fontsize{8}{8}{\thinspace$\pm 0.47$}&$\text{\textcolor{Black}{77.9}}$\fontsize{8}{8}{\thinspace$\pm 0.23$}&$\fontsize{11}{11}{\textbf{\textcolor{Black}{90.9}}}$\fontsize{8}{8}{\thinspace$\pm 0.05$}&$\text{\textcolor{Black}{93.0}}$\fontsize{8}{8}{\thinspace$\pm 0.19$}&$\text{\textcolor{Black}{78.5}}$\fontsize{8}{8}{\thinspace$\pm 0.36$}&$\text{\textcolor{Black}{95.3}}$\fontsize{8}{8}{\thinspace$\pm 0.31$}&$\fontsize{11}{11}{\underline{\textcolor{Black}{56.6}}}$\fontsize{8}{8}{\thinspace$\pm 0.29$}&$\text{\textcolor{Black}{78.1}}$\fontsize{8}{8}{\thinspace$\pm 0.25$}&${\textcolor{Black}{73.9}}$
\\ 
DMN \scriptsize{(CVPR '24)} &$\text{\textcolor{Black}{75.8}}$\fontsize{8}{8}{\thinspace$\pm 0.09$}&$\text{\textcolor{Black}{71.5}}$\fontsize{8}{8}{\thinspace$\pm 0.17$}&$\fontsize{11}{11}{\underline{\textcolor{Black}{31.9}}}$\fontsize{8}{8}{\thinspace$\pm 0.38$}&$\fontsize{11}{11}{\textbf{\textcolor{Black}{64.6}}}$\fontsize{8}{8}{\thinspace$\pm 0.95$}&$\fontsize{11}{11}{\underline{\textcolor{Black}{78.8}}}$\fontsize{8}{8}{\thinspace$\pm 0.29$}&$\text{\textcolor{Black}{90.0}}$\fontsize{8}{8}{\thinspace$\pm 0.08$}&$\fontsize{11}{11}{\underline{\textcolor{Black}{93.4}}}$\fontsize{8}{8}{\thinspace$\pm 0.23$}&$\fontsize{11}{11}{\textbf{\textcolor{Black}{80.9}}}$\fontsize{8}{8}{\thinspace$\pm 0.27$}&$\fontsize{11}{11}{\underline{\textcolor{Black}{95.6}}}$\fontsize{8}{8}{\thinspace$\pm 0.35$}&$\text{\textcolor{Black}{56.1}}$\fontsize{8}{8}{\thinspace$\pm 0.60$}&$\fontsize{11}{11}{\textbf{\textcolor{Black}{79.3}}}$\fontsize{8}{8}{\thinspace$\pm 0.39$}&${\textcolor{Black}{74.4}}$
\\ 
\cellcolor{LightBlue}OGA (ours) &\cellcolor{LightBlue} $\fontsize{11}{11}{\textbf{\textcolor{Black}{76.3}}}$\fontsize{8}{8}{\thinspace$\pm 0.11$}&\cellcolor{LightBlue} $\fontsize{11}{11}{\textbf{\textcolor{Black}{72.2}}}$\fontsize{8}{8}{\thinspace$\pm 0.19$}&\cellcolor{LightBlue} $\fontsize{11}{11}{\textbf{\textcolor{Black}{32.4}}}$\fontsize{8}{8}{\thinspace$\pm 0.40$}&\cellcolor{LightBlue} $\fontsize{11}{11}{\underline{\textcolor{Black}{64.3}}}$\fontsize{8}{8}{\thinspace$\pm 1.02$}&\cellcolor{LightBlue} $\fontsize{11}{11}{\textbf{\textcolor{Black}{79.5}}}$\fontsize{8}{8}{\thinspace$\pm 0.22$}&\cellcolor{LightBlue} $\fontsize{11}{11}{\underline{\textcolor{Black}{90.8}}}$\fontsize{8}{8}{\thinspace$\pm 0.08$}&\cellcolor{LightBlue} $\fontsize{11}{11}{\textbf{\textcolor{Black}{93.9}}}$\fontsize{8}{8}{\thinspace$\pm 0.24$}&\cellcolor{LightBlue} $\fontsize{11}{11}{\underline{\textcolor{Black}{79.9}}}$\fontsize{8}{8}{\thinspace$\pm 0.43$}&\cellcolor{LightBlue} $\fontsize{11}{11}{\textbf{\textcolor{Black}{95.7}}}$\fontsize{8}{8}{\thinspace$\pm 0.33$}&\cellcolor{LightBlue} $\fontsize{11}{11}{\textbf{\textcolor{Black}{57.5}}}$\fontsize{8}{8}{\thinspace$\pm 0.55$}&\cellcolor{LightBlue} $\fontsize{11}{11}{\underline{\textcolor{Black}{79.2}}}$\fontsize{8}{8}{\thinspace$\pm 0.41$}&\cellcolor{LightBlue} $\textbf{\textcolor{Black}{74.7}}$
\end{tabular}}
\end{subtable}
\end{table*}

\vspace{2cm}
\subsection{Results with CNNs architectures.}
\begin{table*}[!h]
\centering
\caption{We show results obtained with CNNs-based architectures and the prompts of Table \ref{tab:prompts}. }

\label{tab:resnets}

\begin{subtable}{\linewidth}

\setlength\dashlinedash{0.2pt}
                        \setlength\dashlinegap{1.5pt}
                        \setlength\arrayrulewidth{0.3pt}
                        \renewcommand{\arraystretch}{1.2}

                                                    \caption{
        With ResNet50.}
    \label{tab:resnet50}
    \centering
    \resizebox{\textwidth}{!}{
        \begin{tabular}{l|cccccccccccc}
          \toprule
 & \rotatebox[origin=c]{45}{ImageNet} & \rotatebox[origin=c]{45}{SUN397} & \rotatebox[origin=c]{45}{Aircraft} & \rotatebox[origin=c]{45}{EuroSAT} & \rotatebox[origin=c]{45}{StanfordCars} & \rotatebox[origin=c]{45}{Food101} & \rotatebox[origin=c]{45}{Pets} & \rotatebox[origin=c]{45}{Flower102} & \rotatebox[origin=c]{45}{Caltech101} & \rotatebox[origin=c]{45}{DTD} & \rotatebox[origin=c]{45}{UCF101} & \rotatebox[origin=c]{45}{Average}
\\ \midrule
\textbf{Zero-Shot} & $\text{58.18}$ & $\text{58.84}$ & $\text{16.95}$ & $\text{36.10}$ & $\text{55.80}$ & $\text{77.36}$ & $\text{85.72}$ & $\text{65.98}$ & $\text{85.92}$ & $\text{42.79}$ & $\text{61.86}$ & $\text{58.7}$
\\ \midrule
TDA \scriptsize{(CVPR '24)} &$\fontsize{11}{11}{\textbf{\textcolor{Black}{59.1}}}$\fontsize{8}{8}{\thinspace$\pm 0.07$}&$\fontsize{11}{11}{\underline{\textcolor{Black}{60.3}}}$\fontsize{8}{8}{\thinspace$\pm 0.13$}&$\fontsize{11}{11}{\underline{\textcolor{Black}{16.2}}}$\fontsize{8}{8}{\thinspace$\pm 0.37$}&$\text{\textcolor{Black}{39.1}}$\fontsize{8}{8}{\thinspace$\pm 1.84$}&$\fontsize{11}{11}{\underline{\textcolor{Black}{56.5}}}$\fontsize{8}{8}{\thinspace$\pm 0.22$}&$\fontsize{11}{11}{\textbf{\textcolor{Black}{77.0}}}$\fontsize{8}{8}{\thinspace$\pm 0.08$}&$\fontsize{11}{11}{\underline{\textcolor{Black}{85.1}}}$\fontsize{8}{8}{\thinspace$\pm 0.30$}&$\fontsize{11}{11}{\textbf{\textcolor{Black}{67.1}}}$\fontsize{8}{8}{\thinspace$\pm 0.34$}&$\fontsize{11}{11}{\textbf{\textcolor{Black}{86.9}}}$\fontsize{8}{8}{\thinspace$\pm 0.44$}&$\fontsize{11}{11}{\underline{\textcolor{Black}{42.8}}}$\fontsize{8}{8}{\thinspace$\pm 0.35$}&$\fontsize{11}{11}{\underline{\textcolor{Black}{62.7}}}$\fontsize{8}{8}{\thinspace$\pm 0.33$}&$\underline{\textcolor{Black}{59.3}}$
\\ 
DMN \scriptsize{(CVPR '24)} &$\text{\textcolor{Black}{57.2}}$\fontsize{8}{8}{\thinspace$\pm 0.10$}&$\text{\textcolor{Black}{59.2}}$\fontsize{8}{8}{\thinspace$\pm 0.18$}&$\text{\textcolor{Black}{15.8}}$\fontsize{8}{8}{\thinspace$\pm 0.33$}&$\fontsize{11}{11}{\textbf{\textcolor{Black}{44.8}}}$\fontsize{8}{8}{\thinspace$\pm 1.91$}&$\text{\textcolor{Black}{55.3}}$\fontsize{8}{8}{\thinspace$\pm 0.34$}&$\text{\textcolor{Black}{73.6}}$\fontsize{8}{8}{\thinspace$\pm 0.13$}&$\text{\textcolor{Black}{83.3}}$\fontsize{8}{8}{\thinspace$\pm 0.43$}&$\fontsize{11}{11}{\underline{\textcolor{Black}{66.5}}}$\fontsize{8}{8}{\thinspace$\pm 0.37$}&$\text{\textcolor{Black}{85.3}}$\fontsize{8}{8}{\thinspace$\pm 0.59$}&$\text{\textcolor{Black}{42.2}}$\fontsize{8}{8}{\thinspace$\pm 0.60$}&$\text{\textcolor{Black}{61.9}}$\fontsize{8}{8}{\thinspace$\pm 0.46$}&${\textcolor{Black}{58.6}}$
\\ 
\cellcolor{LightBlue} OGA (ours) &\cellcolor{LightBlue} $\fontsize{11}{11}{\underline{\textcolor{Black}{58.8}}}$\fontsize{8}{8}{\thinspace$\pm 0.12$}&\cellcolor{LightBlue} $\fontsize{11}{11}{\textbf{\textcolor{Black}{61.3}}}$\fontsize{8}{8}{\thinspace$\pm 0.14$}&\cellcolor{LightBlue} $\fontsize{11}{11}{\textbf{\textcolor{Black}{16.3}}}$\fontsize{8}{8}{\thinspace$\pm 0.34$}&\cellcolor{LightBlue} $\fontsize{11}{11}{\underline{\textcolor{Black}{43.8}}}$\fontsize{8}{8}{\thinspace$\pm 1.97$}&\cellcolor{LightBlue} $\fontsize{11}{11}{\textbf{\textcolor{Black}{57.7}}}$\fontsize{8}{8}{\thinspace$\pm 0.22$}&\cellcolor{LightBlue} $\fontsize{11}{11}{\underline{\textcolor{Black}{76.1}}}$\fontsize{8}{8}{\thinspace$\pm 0.13$}&\cellcolor{LightBlue} $\fontsize{11}{11}{\textbf{\textcolor{Black}{85.5}}}$\fontsize{8}{8}{\thinspace$\pm 0.37$}&\cellcolor{LightBlue} $\text{\textcolor{Black}{66.1}}$\fontsize{8}{8}{\thinspace$\pm 0.44$}&\cellcolor{LightBlue} $\fontsize{11}{11}{\underline{\textcolor{Black}{85.4}}}$\fontsize{8}{8}{\thinspace$\pm 0.58$}&\cellcolor{LightBlue} $\fontsize{11}{11}{\textbf{\textcolor{Black}{43.9}}}$\fontsize{8}{8}{\thinspace$\pm 0.55$}&\cellcolor{LightBlue} $\fontsize{11}{11}{\textbf{\textcolor{Black}{62.9}}}$\fontsize{8}{8}{\thinspace$\pm 0.44$}&\cellcolor{LightBlue} $\textbf{\textcolor{Black}{59.8}}$

\end{tabular}}
\end{subtable}

\vspace{0.1cm}

\begin{subtable}{\linewidth}
\setlength\dashlinedash{0.2pt}
                        \setlength\dashlinegap{1.5pt}
                        \setlength\arrayrulewidth{0.3pt}
                        \renewcommand{\arraystretch}{1.2}
                            \caption{
        With ResNet101.}
    \label{tab:resnet101}
    \centering
    \resizebox{\textwidth}{!}{
        \begin{tabular}{l|cccccccccccc}
          \toprule
 & \rotatebox[origin=c]{45}{ImageNet} & \rotatebox[origin=c]{45}{SUN397} & \rotatebox[origin=c]{45}{Aircraft} & \rotatebox[origin=c]{45}{EuroSAT} & \rotatebox[origin=c]{45}{StanfordCars} & \rotatebox[origin=c]{45}{Food101} & \rotatebox[origin=c]{45}{Pets} & \rotatebox[origin=c]{45}{Flower102} & \rotatebox[origin=c]{45}{Caltech101} & \rotatebox[origin=c]{45}{DTD} & \rotatebox[origin=c]{45}{UCF101} & \rotatebox[origin=c]{45}{Average}
\\ \midrule
\textbf{Zero-Shot} & $\text{61.26}$ & $\text{59.04}$ & $\text{18.12}$ & $\text{32.80}$ & $\text{63.15}$ & $\text{80.67}$ & $\text{86.89}$ & $\text{64.35}$ & $\text{90.02}$ & $\text{37.06}$ & $\text{61.01}$ & $\text{59.5}$
\\ \midrule
TDA \scriptsize{(CVPR '24)} &$\fontsize{11}{11}{\underline{\textcolor{Black}{62.4}}}$\fontsize{8}{8}{\thinspace$\pm 0.07$}&$\text{\textcolor{Black}{60.7}}$\fontsize{8}{8}{\thinspace$\pm 0.14$}&$\fontsize{11}{11}{\underline{\textcolor{Black}{17.8}}}$\fontsize{8}{8}{\thinspace$\pm 0.34$}&$\text{\textcolor{Black}{41.2}}$\fontsize{8}{8}{\thinspace$\pm 0.70$}&$\text{\textcolor{Black}{63.5}}$\fontsize{8}{8}{\thinspace$\pm 0.22$}&$\fontsize{11}{11}{\underline{\textcolor{Black}{80.4}}}$\fontsize{8}{8}{\thinspace$\pm 0.08$}&$\text{\textcolor{Black}{86.2}}$\fontsize{8}{8}{\thinspace$\pm 0.25$}&$\text{\textcolor{Black}{64.4}}$\fontsize{8}{8}{\thinspace$\pm 0.41$}&$\fontsize{11}{11}{\textbf{\textcolor{Black}{89.5}}}$\fontsize{8}{8}{\thinspace$\pm 0.47$}&$\text{\textcolor{Black}{38.1}}$\fontsize{8}{8}{\thinspace$\pm 0.42$}&$\text{\textcolor{Black}{62.6}}$\fontsize{8}{8}{\thinspace$\pm 0.31$}&${\textcolor{Black}{60.6}}$
\\ 
DMN \scriptsize{(CVPR '24)} &$\text{\textcolor{Black}{62.2}}$\fontsize{8}{8}{\thinspace$\pm 0.10$}&$\fontsize{11}{11}{\underline{\textcolor{Black}{61.4}}}$\fontsize{8}{8}{\thinspace$\pm 0.18$}&$\text{\textcolor{Black}{17.5}}$\fontsize{8}{8}{\thinspace$\pm 0.34$}&$\fontsize{11}{11}{\underline{\textcolor{Black}{41.5}}}$\fontsize{8}{8}{\thinspace$\pm 1.03$}&$\fontsize{11}{11}{\underline{\textcolor{Black}{64.2}}}$\fontsize{8}{8}{\thinspace$\pm 0.30$}&$\text{\textcolor{Black}{79.3}}$\fontsize{8}{8}{\thinspace$\pm 0.11$}&$\fontsize{11}{11}{\underline{\textcolor{Black}{87.0}}}$\fontsize{8}{8}{\thinspace$\pm 0.32$}&$\fontsize{11}{11}{\textbf{\textcolor{Black}{66.4}}}$\fontsize{8}{8}{\thinspace$\pm 0.34$}&$\text{\textcolor{Black}{89.1}}$\fontsize{8}{8}{\thinspace$\pm 0.50$}&$\fontsize{11}{11}{\underline{\textcolor{Black}{38.4}}}$\fontsize{8}{8}{\thinspace$\pm 0.58$}&$\fontsize{11}{11}{\underline{\textcolor{Black}{64.0}}}$\fontsize{8}{8}{\thinspace$\pm 0.49$}&$\underline{\textcolor{Black}{61.0}}$
\\ 
\cellcolor{LightBlue} OGA (ours) &\cellcolor{LightBlue} $\fontsize{11}{11}{\textbf{\textcolor{Black}{62.6}}}$\fontsize{8}{8}{\thinspace$\pm 0.10$}&\cellcolor{LightBlue} $\fontsize{11}{11}{\textbf{\textcolor{Black}{61.9}}}$\fontsize{8}{8}{\thinspace$\pm 0.15$}&\cellcolor{LightBlue} $\fontsize{11}{11}{\textbf{\textcolor{Black}{17.9}}}$\fontsize{8}{8}{\thinspace$\pm 0.31$}&\cellcolor{LightBlue} $\fontsize{11}{11}{\textbf{\textcolor{Black}{44.4}}}$\fontsize{8}{8}{\thinspace$\pm 1.22$}&\cellcolor{LightBlue} $\fontsize{11}{11}{\textbf{\textcolor{Black}{64.5}}}$\fontsize{8}{8}{\thinspace$\pm 0.21$}&\cellcolor{LightBlue} $\fontsize{11}{11}{\textbf{\textcolor{Black}{80.6}}}$\fontsize{8}{8}{\thinspace$\pm 0.11$}&\cellcolor{LightBlue} $\fontsize{11}{11}{\textbf{\textcolor{Black}{87.6}}}$\fontsize{8}{8}{\thinspace$\pm 0.26$}&\cellcolor{LightBlue} $\fontsize{11}{11}{\underline{\textcolor{Black}{65.4}}}$\fontsize{8}{8}{\thinspace$\pm 0.34$}&\cellcolor{LightBlue} $\fontsize{11}{11}{\underline{\textcolor{Black}{89.2}}}$\fontsize{8}{8}{\thinspace$\pm 0.48$}&\cellcolor{LightBlue} $\fontsize{11}{11}{\textbf{\textcolor{Black}{39.3}}}$\fontsize{8}{8}{\thinspace$\pm 0.59$}&\cellcolor{LightBlue} $\fontsize{11}{11}{\textbf{\textcolor{Black}{64.7}}}$\fontsize{8}{8}{\thinspace$\pm 0.38$}&\cellcolor{LightBlue} $\textbf{\textcolor{Black}{61.6}}$
\end{tabular}}
\end{subtable}

\end{table*}
\begin{table*}[!h]
\centering
\caption{We show results obtained with CNNs-based architectures and the custom ensemble of prompts of Table \ref{tab:ct}. }

\label{tab:resnets_ensemble}

\begin{subtable}{\linewidth}

\setlength\dashlinedash{0.2pt}
                        \setlength\dashlinegap{1.5pt}
                        \setlength\arrayrulewidth{0.3pt}
                        \renewcommand{\arraystretch}{1.2}

                                                    \caption{
        With ResNet50.}
    \label{tab:resnet50_ensemble}
        \centering
    \resizebox{\textwidth}{!}{
        \begin{tabular}{l|cccccccccccc}
          \toprule
 & \rotatebox[origin=c]{45}{ImageNet} & \rotatebox[origin=c]{45}{SUN397} & \rotatebox[origin=c]{45}{Aircraft} & \rotatebox[origin=c]{45}{EuroSAT} & \rotatebox[origin=c]{45}{StanfordCars} & \rotatebox[origin=c]{45}{Food101} & \rotatebox[origin=c]{45}{Pets} & \rotatebox[origin=c]{45}{Flower102} & \rotatebox[origin=c]{45}{Caltech101} & \rotatebox[origin=c]{45}{DTD} & \rotatebox[origin=c]{45}{UCF101} & \rotatebox[origin=c]{45}{Average}
\\ \midrule
\textbf{Zero-Shot} & $\text{60.25}$ & $\text{60.96}$ & $\text{16.41}$ & $\text{27.09}$ & $\text{56.31}$ & $\text{76.42}$ & $\text{82.77}$ & $\text{62.65}$ & $\text{87.79}$ & $\text{40.48}$ & $\text{60.16}$ & $\text{57.4}$
\\ \midrule
TDA \scriptsize{(CVPR '24)} &$\fontsize{11}{11}{\textbf{\textcolor{Black}{60.7}}}$\fontsize{8}{8}{\thinspace$\pm 0.06$}&$\fontsize{11}{11}{\underline{\textcolor{Black}{62.0}}}$\fontsize{8}{8}{\thinspace$\pm 0.11$}&$\fontsize{11}{11}{\underline{\textcolor{Black}{15.6}}}$\fontsize{8}{8}{\thinspace$\pm 0.33$}&$\text{\textcolor{Black}{31.8}}$\fontsize{8}{8}{\thinspace$\pm 1.52$}&$\fontsize{11}{11}{\underline{\textcolor{Black}{56.9}}}$\fontsize{8}{8}{\thinspace$\pm 0.24$}&$\fontsize{11}{11}{\textbf{\textcolor{Black}{75.9}}}$\fontsize{8}{8}{\thinspace$\pm 0.09$}&$\fontsize{11}{11}{\underline{\textcolor{Black}{82.7}}}$\fontsize{8}{8}{\thinspace$\pm 0.28$}&$\fontsize{11}{11}{\underline{\textcolor{Black}{64.1}}}$\fontsize{8}{8}{\thinspace$\pm 0.41$}&$\fontsize{11}{11}{\textbf{\textcolor{Black}{88.2}}}$\fontsize{8}{8}{\thinspace$\pm 0.43$}&$\fontsize{11}{11}{\underline{\textcolor{Black}{39.8}}}$\fontsize{8}{8}{\thinspace$\pm 0.42$}&$\fontsize{11}{11}{\underline{\textcolor{Black}{61.6}}}$\fontsize{8}{8}{\thinspace$\pm 0.29$}&$\underline{\textcolor{Black}{58.1}}$
\\ 
DMN \scriptsize{(CVPR '24)} &$\text{\textcolor{Black}{58.4}}$\fontsize{8}{8}{\thinspace$\pm 0.10$}&$\text{\textcolor{Black}{60.5}}$\fontsize{8}{8}{\thinspace$\pm 0.17$}&$\text{\textcolor{Black}{15.2}}$\fontsize{8}{8}{\thinspace$\pm 0.30$}&$\fontsize{11}{11}{\underline{\textcolor{Black}{33.1}}}$\fontsize{8}{8}{\thinspace$\pm 1.43$}&$\text{\textcolor{Black}{55.9}}$\fontsize{8}{8}{\thinspace$\pm 0.35$}&$\text{\textcolor{Black}{72.8}}$\fontsize{8}{8}{\thinspace$\pm 0.15$}&$\text{\textcolor{Black}{81.3}}$\fontsize{8}{8}{\thinspace$\pm 0.37$}&$\fontsize{11}{11}{\textbf{\textcolor{Black}{64.5}}}$\fontsize{8}{8}{\thinspace$\pm 0.39$}&$\fontsize{11}{11}{\underline{\textcolor{Black}{87.3}}}$\fontsize{8}{8}{\thinspace$\pm 0.55$}&$\text{\textcolor{Black}{39.3}}$\fontsize{8}{8}{\thinspace$\pm 0.64$}&$\text{\textcolor{Black}{61.0}}$\fontsize{8}{8}{\thinspace$\pm 0.51$}&${\textcolor{Black}{57.2}}$
\\ 
\cellcolor{LightBlue} OGA (ours) &\cellcolor{LightBlue} $\fontsize{11}{11}{\underline{\textcolor{Black}{59.7}}}$\fontsize{8}{8}{\thinspace$\pm 0.12$}&\cellcolor{LightBlue} $\fontsize{11}{11}{\textbf{\textcolor{Black}{63.0}}}$\fontsize{8}{8}{\thinspace$\pm 0.13$}&\cellcolor{LightBlue} $\fontsize{11}{11}{\textbf{\textcolor{Black}{15.8}}}$\fontsize{8}{8}{\thinspace$\pm 0.33$}&\cellcolor{LightBlue} $\fontsize{11}{11}{\textbf{\textcolor{Black}{34.3}}}$\fontsize{8}{8}{\thinspace$\pm 1.56$}&\cellcolor{LightBlue} $\fontsize{11}{11}{\textbf{\textcolor{Black}{58.2}}}$\fontsize{8}{8}{\thinspace$\pm 0.24$}&\cellcolor{LightBlue} $\fontsize{11}{11}{\underline{\textcolor{Black}{75.3}}}$\fontsize{8}{8}{\thinspace$\pm 0.15$}&\cellcolor{LightBlue} $\fontsize{11}{11}{\textbf{\textcolor{Black}{83.3}}}$\fontsize{8}{8}{\thinspace$\pm 0.34$}&\cellcolor{LightBlue} $\text{\textcolor{Black}{63.5}}$\fontsize{8}{8}{\thinspace$\pm 0.44$}&\cellcolor{LightBlue} $\text{\textcolor{Black}{87.0}}$\fontsize{8}{8}{\thinspace$\pm 0.55$}&\cellcolor{LightBlue} $\fontsize{11}{11}{\textbf{\textcolor{Black}{40.3}}}$\fontsize{8}{8}{\thinspace$\pm 0.58$}&\cellcolor{LightBlue} $\fontsize{11}{11}{\textbf{\textcolor{Black}{61.8}}}$\fontsize{8}{8}{\thinspace$\pm 0.49$}&\cellcolor{LightBlue} $\textbf{\textcolor{Black}{58.4}}$
\end{tabular}}
\end{subtable}

\vspace{0.1cm}

\begin{subtable}{\linewidth}
\setlength\dashlinedash{0.2pt}
                        \setlength\dashlinegap{1.5pt}
                        \setlength\arrayrulewidth{0.3pt}
                        \renewcommand{\arraystretch}{1.2}
                            \caption{
        With ResNet101.}
    \label{tab:resnet101_ensemble}
     \centering
    \resizebox{\textwidth}{!}{
        \begin{tabular}{l|cccccccccccc}
          \toprule
 & \rotatebox[origin=c]{45}{ImageNet} & \rotatebox[origin=c]{45}{SUN397} & \rotatebox[origin=c]{45}{Aircraft} & \rotatebox[origin=c]{45}{EuroSAT} & \rotatebox[origin=c]{45}{StanfordCars} & \rotatebox[origin=c]{45}{Food101} & \rotatebox[origin=c]{45}{Pets} & \rotatebox[origin=c]{45}{Flower102} & \rotatebox[origin=c]{45}{Caltech101} & \rotatebox[origin=c]{45}{DTD} & \rotatebox[origin=c]{45}{UCF101} & \rotatebox[origin=c]{45}{Average}
\\ \midrule
\textbf{Zero-Shot} & $\text{62.46}$ & $\text{61.06}$ & $\text{17.61}$ & $\text{25.09}$ & $\text{62.88}$ & $\text{80.68}$ & $\text{84.79}$ & $\text{61.88}$ & $\text{90.83}$ & $\text{41.49}$ & $\text{60.69}$ & $\text{59.0}$
\\ \midrule
TDA \scriptsize{(CVPR '24)} &$\fontsize{11}{11}{\underline{\textcolor{Black}{63.0}}}$\fontsize{8}{8}{\thinspace$\pm 0.08$}&$\text{\textcolor{Black}{62.1}}$\fontsize{8}{8}{\thinspace$\pm 0.12$}&$\fontsize{11}{11}{\underline{\textcolor{Black}{16.9}}}$\fontsize{8}{8}{\thinspace$\pm 0.30$}&$\text{\textcolor{Black}{28.6}}$\fontsize{8}{8}{\thinspace$\pm 1.02$}&$\text{\textcolor{Black}{63.5}}$\fontsize{8}{8}{\thinspace$\pm 0.24$}&$\fontsize{11}{11}{\underline{\textcolor{Black}{80.1}}}$\fontsize{8}{8}{\thinspace$\pm 0.08$}&$\text{\textcolor{Black}{84.5}}$\fontsize{8}{8}{\thinspace$\pm 0.26$}&$\text{\textcolor{Black}{61.8}}$\fontsize{8}{8}{\thinspace$\pm 0.45$}&$\fontsize{11}{11}{\textbf{\textcolor{Black}{90.0}}}$\fontsize{8}{8}{\thinspace$\pm 0.47$}&$\text{\textcolor{Black}{40.9}}$\fontsize{8}{8}{\thinspace$\pm 0.45$}&$\text{\textcolor{Black}{62.1}}$\fontsize{8}{8}{\thinspace$\pm 0.34$}&${\textcolor{Black}{59.4}}$
\\ 
DMN \scriptsize{(CVPR '24)} &$\text{\textcolor{Black}{62.9}}$\fontsize{8}{8}{\thinspace$\pm 0.10$}&$\fontsize{11}{11}{\underline{\textcolor{Black}{62.6}}}$\fontsize{8}{8}{\thinspace$\pm 0.16$}&$\text{\textcolor{Black}{16.8}}$\fontsize{8}{8}{\thinspace$\pm 0.32$}&$\fontsize{11}{11}{\underline{\textcolor{Black}{33.7}}}$\fontsize{8}{8}{\thinspace$\pm 1.51$}&$\fontsize{11}{11}{\textbf{\textcolor{Black}{64.8}}}$\fontsize{8}{8}{\thinspace$\pm 0.29$}&$\text{\textcolor{Black}{79.3}}$\fontsize{8}{8}{\thinspace$\pm 0.10$}&$\fontsize{11}{11}{\underline{\textcolor{Black}{85.1}}}$\fontsize{8}{8}{\thinspace$\pm 0.28$}&$\fontsize{11}{11}{\textbf{\textcolor{Black}{64.3}}}$\fontsize{8}{8}{\thinspace$\pm 0.44$}&$\text{\textcolor{Black}{89.4}}$\fontsize{8}{8}{\thinspace$\pm 0.52$}&$\fontsize{11}{11}{\underline{\textcolor{Black}{41.1}}}$\fontsize{8}{8}{\thinspace$\pm 0.54$}&$\fontsize{11}{11}{\underline{\textcolor{Black}{63.4}}}$\fontsize{8}{8}{\thinspace$\pm 0.43$}&$\underline{\textcolor{Black}{60.3}}$
\\ 
\cellcolor{LightBlue} OGA (ours) &\cellcolor{LightBlue} $\fontsize{11}{11}{\textbf{\textcolor{Black}{63.0}}}$\fontsize{8}{8}{\thinspace$\pm 0.11$}&\cellcolor{LightBlue} $\fontsize{11}{11}{\textbf{\textcolor{Black}{62.6}}}$\fontsize{8}{8}{\thinspace$\pm 0.18$}&\cellcolor{LightBlue} $\fontsize{11}{11}{\textbf{\textcolor{Black}{17.2}}}$\fontsize{8}{8}{\thinspace$\pm 0.34$}&\cellcolor{LightBlue} $\fontsize{11}{11}{\textbf{\textcolor{Black}{34.4}}}$\fontsize{8}{8}{\thinspace$\pm 1.38$}&\cellcolor{LightBlue} $\fontsize{11}{11}{\underline{\textcolor{Black}{64.5}}}$\fontsize{8}{8}{\thinspace$\pm 0.20$}&\cellcolor{LightBlue} $\fontsize{11}{11}{\textbf{\textcolor{Black}{80.5}}}$\fontsize{8}{8}{\thinspace$\pm 0.10$}&\cellcolor{LightBlue} $\fontsize{11}{11}{\textbf{\textcolor{Black}{86.3}}}$\fontsize{8}{8}{\thinspace$\pm 0.27$}&\cellcolor{LightBlue} $\fontsize{11}{11}{\underline{\textcolor{Black}{63.0}}}$\fontsize{8}{8}{\thinspace$\pm 0.31$}&\cellcolor{LightBlue} $\fontsize{11}{11}{\underline{\textcolor{Black}{89.6}}}$\fontsize{8}{8}{\thinspace$\pm 0.46$}&\cellcolor{LightBlue} $\fontsize{11}{11}{\textbf{\textcolor{Black}{41.7}}}$\fontsize{8}{8}{\thinspace$\pm 0.47$}&\cellcolor{LightBlue} $\fontsize{11}{11}{\textbf{\textcolor{Black}{63.5}}}$\fontsize{8}{8}{\thinspace$\pm 0.38$}&\cellcolor{LightBlue} $\textbf{\textcolor{Black}{60.6}}$

\end{tabular}}
\end{subtable}

\end{table*}

\subsection{Summary.}
\begin{table*}[h]
\centering
\caption{We show results averaged over the 11 datasets. Standard prompts refer to Table~\ref{tab:prompts} while Custom Ensemble corresponds to Table~\ref{tab:ct}.}

\label{tab:summary_backbones}

\setlength\dashlinedash{0.2pt}
                        \setlength\dashlinegap{1.5pt}
                        \setlength\arrayrulewidth{0.3pt}
                        \renewcommand{\arraystretch}{1.2}
    \centering
        \begin{tabular}{ll|ccccc}
          \toprule
& & ViT-B/16 & ViT-B/32 & ViT-L/14 & ResNet50 & ResNet101 \\ \midrule

\multirow{4}{*}{\rotatebox[origin=c]{90}{\scriptsize{Standard Prompts}}} & \textbf{Zero-Shot} & 65.3 & 61.9 & 72.6 & 58.7 & 59.5 \\ \cmidrule[0.5pt](lr{0em}){2-7}
& TDA & \underline{67.7} & \underline{62.3} & 73.5 & \underline{59.3} & 60.6 \\
& DMN & 67.5 & 61.8 & \underline{73.7} & 58.6 & \underline{61.0} \\
& \cellcolor{LightBlue}OGA (ours) & \cellcolor{LightBlue}\textbf{68.5} & \cellcolor{LightBlue}\textbf{62.9} & \cellcolor{LightBlue}\textbf{74.3} & \cellcolor{LightBlue}\textbf{59.8} & \cellcolor{LightBlue}\textbf{61.6} \\ \midrule

\multirow{4}{*}{\rotatebox[origin=c]{90}{\scriptsize{Custom Ensemble}}} & \textbf{Zero-Shot} & 65.6 & 61.4 & 72.2 & 57.4 & 59.0\\ \cmidrule[0.5pt](lr{0em}){2-7}
& TDA & \underline{66.9} & \underline{62.3} & 73.9 & \underline{58.1} & 59.4 \\
& DMN & 66.4 & 61.6 & \underline{74.4} & 57.2 & \underline{60.3} \\
& \cellcolor{LightBlue}OGA (ours) & \cellcolor{LightBlue}\textbf{67.3} & \cellcolor{LightBlue}\textbf{62.8} & \cellcolor{LightBlue}\textbf{74.7} & \cellcolor{LightBlue}\textbf{58.4} & \cellcolor{LightBlue}\textbf{60.6} \\


\end{tabular}

\end{table*}

\section{Hyper-Parameters.}
Both comparative methods use per-dataset hyper-parameters in their benchmarks. Since we do not have access to ground truth labels to tune those hyper-parameters in a TTA scenario, for a more rigorous comparison we use the same \textit{fixed hyper-parameters} for all datasets, i.e. the ones they tuned for ImageNet. 
For TDA, this means the positive logits mixing coefficients is set to $2$, while the negative logits mixing coefficient is set to $0.117$.
For DMN, since we only consider zero-shot scenarios, we only need to set the coefficient relative to the dynamic memory, which is therefore kept fixed at $1$. As highlighted in the main paper, the hyper-parameter $\nu$ of OGA is always fixed at $0.05$.
\end{document}